\documentclass{article}

\PassOptionsToPackage{numbers, compress}{natbib}

\usepackage[final]{neurips_2022}

\usepackage[utf8]{inputenc} 
\usepackage[T1]{fontenc}    
\usepackage{hyperref}       
\usepackage{url}            
\usepackage{booktabs}       
\usepackage{amsfonts}       
\usepackage{nicefrac}       
\usepackage{microtype}      
\usepackage{xcolor}         
\usepackage{algorithm}
\usepackage{algpseudocode}
\usepackage{amsmath,amssymb}
\usepackage{graphicx}
\usepackage{subcaption}
\usepackage{enumitem}
\usepackage{multirow}

\DeclareMathOperator{\E}{\mathbb{E}}

\title{Generating multivariate time series with COmmon Source CoordInated GAN (COSCI-GAN)}

\author{
  Ali Seyfi \\
  Department of Computer Science\\
  University of British Columbia\\
  Vancouver, BC \\
  \texttt{aliseyfi@cs.ubc.ca} \\
  \And
  Jean-Francois Rajotte \\
  Data Science Institute \\
  University of British Columbia \\
  Vancouver, BC \\
  \texttt{jfraj@mail.ubc.ca} \\
  \And
  Raymond T. Ng \\
  Department of Computer Science \\
  University of British Columbia \\
  Vancouver, BC \\
  \texttt{rng@cs.ubc.ca} \\
}

\begin{document}

\maketitle

\begin{abstract}
Generating multivariate time series is a promising approach for sharing sensitive data in many medical, financial, and IoT applications. A common type of multivariate time series originates from a single source such as the biometric measurements from a medical patient. This leads to complex dynamical patterns between individual time series that are hard to learn by typical generation models such as GANs. There is valuable information in those patterns that machine learning models can use to better classify, predict or perform other downstream tasks. We propose a novel framework that takes time series' common origin into account and favors channel/feature relationships preservation. The two key points of our method are: 1) the individual time series are generated from a common point in latent space and 2) a central discriminator favors the preservation of inter-channel/feature dynamics. We demonstrate empirically that our method helps preserve channel/feature correlations and that our synthetic data performs very well in downstream tasks with medical and financial data.
\end{abstract}

\section{Introduction}
\label{Introduction}

Multivariate Time Series (MTS) are composed of individual time series (TS) sharing the same time reference. In some cases, the individual time series further share a common source such as the biometric values from a medical patient, the stock prices from economic events or geographically separated seismic measurements from a single earthquake. This leads to specific correlation patterns and time dynamics across the time series. Such complex patterns can be crucial when a model is trained on MTS and might need a huge amount of training samples to be captured by a machine learning algorithm. For many applications, however, there may not be enough high-quality training data. For example, for many biomedical and health care applications, data scarcity is common and data sharing to build a larger training set is challenging due to regulatory requirements or ethical concerns (\cite{hipaa}, \cite{gdpr}).
Such concerns are justified; it is well known that sharing data associated with a single individual, even anonymized, can lead to unexpected privacy breaches (\cite{Sweeney2000SimpleDO}, \cite{Mandl2021HIPAAAT}, \cite{4531148}, and \cite{9064731}).
Synthetic MTS could be an attractive alternative to share the patterns and statistical information of an MTS dataset.
If done properly, synthetic MTS should not have a one-to-one mapping to the original data, although it can come with its own privacy and quality challenges (see \cite{Stadler2020SyntheticD} for example). 

Beyond data sharing, synthetic MTS can be used for augmenting a dataset to improve the performance of a trained model \cite{Nalepa2019DataAF} or increase the contribution of underrepresented sub-populations \cite{Rajotte2021ReducingBA}.
For example, in the health domain, a patient can generate multiple TS from biometric measurements, wearable and IoT sensors, but the collection of such data may not have a representative coverage of a full population.
Synthetic MTS can help to augment datasets for improving downstream analysis, such as for forecasting and classification tasks.
In this work, we are interested in generating synthetic MTS that both preserve utility and statistical properties of the original data.
By utility we mean the ability to support a specific downstream task such as the performance of a classifier when trained on synthetic data.
Preserving statistical properties such as channel or feature\footnote{Note that we use channel and feature interchangeably here since the term channel is commonly used for EEG and other medical time series that partly motivate this work. From a typical machine learning context, a channel is basically a feature. Generally in this paper, feature and channel corresponds to a collection of recorded data point from a given sensor.} correlations increase the potential benefits to an unforeseen downstream task, exploratory data analysis or educational purposes when sharing data is not possible. We observe that current methods often struggle at correlation preservation needed for downstream task and our method addresses these limitations.

In this paper, we propose a novel method for generating MTS that comes from a common source by defining an architecture that explicitly takes inter-channel correlations into account. While MTS can be created with a typical generative deep learning architecture (such as VAE or GAN) with multiple channels output, we will demonstrate that our method not only preserves the quality of each individual TS by generating each individual TS separately from a common noise but also preserves the relationships between TS by having a central discriminator receiving all the generated individual TS as a single input. We consider \textit{noise} to be a point of random sampling in the latent space that represents a patient's "whole biological environment." By common noise, we refer to a common point in the latent space, i.e., that latent space is the patient space, and we sample a patient by sampling a noise for the generator input.

We perform extensive empirical evaluation on MTS datasets where individual TS originate from a single source. We evaluate the resulting synthetic MTS by comparing their statistical features with the real MTS and their utility on downstream classification task.
We pick classification to measure the utility of the synthetic data because classification is one of the most popular analytic tasks in machine learning.
We also compare the real and synthetic MTS visually in embedding spaces, and evaluate our method against state-of-the-art baseline methods. Our contributions can be summarized as follows:

\begin{itemize}
\item To our knowledge, this is the first study to analyse how to generate multivariate time series with individual channel generation originating from a common noise while inter-channel correlations preservation is forced with a central discriminator.

\item Demonstrating that COSCI-GAN compares favourably with state-of-the-art algorithms in downstream tasks on an Electroencephalography (EEG) eye state time series dataset.

\item Demonstrating that COSCI-GAN results compares favourably with state-of-the-art algorithms in preservation statistical properties of on the EEG dataset.

\item Open sourcing the implementation of our methods and experiments\footnote{\url{https://github.com/aliseyfi75/COSCI-GAN}}.
\end{itemize}

The rest of the paper is arranged as follows: Section 2 discusses related work. Section 3 formalizes he problem description. Section 4 presents our COSCI-GAN model architecture and gives implementation details. Section 5 presents an extensive empirical evaluation. Section 6 discusses the benefits, limitations and potential negative societal impacts.

\section{Related Work}
\label{Related}

Synthetic data is often proposed to provide or to increase privacy protection \cite{info:doi/10.2196/23139}. The only privacy protection framework with predictable effect on privacy is to incorporate {\em differential privacy (DP)} into the learning algorithm (\cite{Abadi2016DeepLW}) by adding calibrated noise to the parameter updates.
For every non-DP generation method, the privacy must be assessed empirically with privacy attacks (\cite{Carlini2021MembershipIA}).

Generative Adversarial Networks \cite{Goodfellow2014GenerativeAN} are a popular method to generate synthetic data. Since their inception, they have expanded to include time-series data production, see \cite{Brophy2021GenerativeAN} for a comprehensive overview.
Mogren introduced the C-RNN-GAN approach, which employs RNNs as both the generator and discriminator in order to synthesise time series from a random vector \cite{Mogren2016CRNNGANCR}.
Esteban et al. later proposed RCGAN  to create medical data using a similar architecture \cite{Esteban2017RealvaluedT}.
These frameworks have been applied to a wide range of application domains, including biosignals \cite{Harada2018BiosignalDA}, finance \cite{Simonetto2018GeneratingST}, sensor \cite{Alzantot2017SenseGenAD}, text \cite{Zhang2016GeneratingTV}, and smart grid data \cite{Zhang2018GenerativeAN}.
However, the typical framework and loss function of GANs are insufficient for the production of multivariate time series, especially if we want to preserve the correlation among the channels as we will demonstrate below.
Xu et al. developed COT-GAN based on ideas of optimal transport theory \cite{Xu2020COTGANGS} and more recently, Li et al. developed TTS-GAN  which is capable of generating realistic synthetic time series of any length \cite{Li2022TTSGANAT} both work focusing solely on statistical evaluation of the data such as correlations.

Yoon et al. developped TimeGANs based on recurrent conditional GAN for capturing the temporal dynamics of data throughout time \cite{Yoon2019TimeseriesGA}.
It entails training supervised and unsupervised targets concurrently using a learnt embedding space.
It creates time series data by learning an embedding space and optimising it via binary adversarial feedback and stepwise supervised loss.
Most recently, Fourier Flows \cite{Alaa2021GenerativeTM} was proposed as a method based on a Fourier transform layer followed by a chain of spectral filters leading to an exact likelihood optimization.

\section{Problem Formulation}
\label{Problem}

\begin{figure}[htp]
  \centering
  \includegraphics[width=0.8\linewidth]{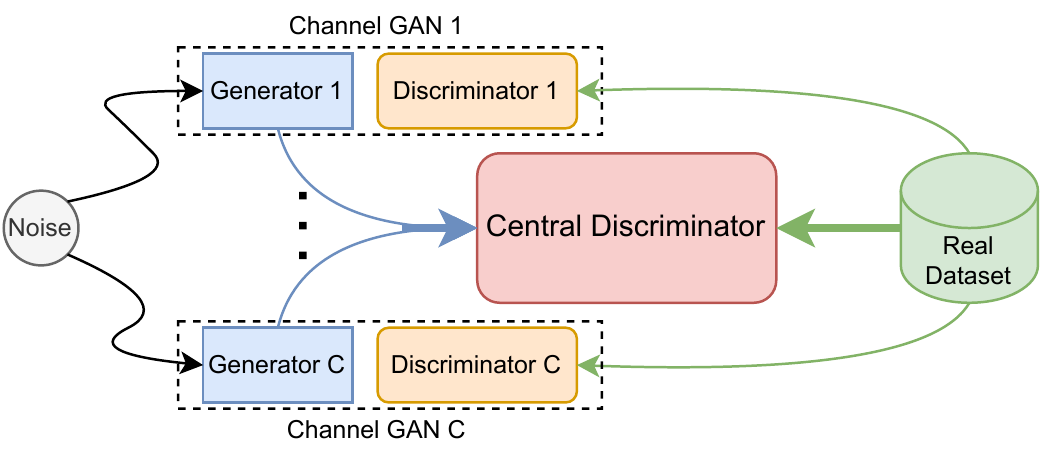}
  \caption{Overall structure of COSCI-GAN. Each Channel GAN is dedicated to one channel/feature with input either a real one or a fake one with their discriminator being a binary classifier.
  The central discriminator, a binary classifier, receives full instances (i.e. all channel) either all real or all fake.
  }
  \label{fig:COSCI-GAN}
  \vspace{-10pt}
\end{figure}

Let $X$ be an MTS dataset of $N$ instances, each composed of $C$ channels $X_i$, where $i \in  {1, \dots, C}$.
An instance of $X$ can be described as $x^n = \{(x^n_1, \dots, x^n_C)\}$.
We want to find a distribution $q(X_1, \dots, X_C)$ that is as close to $p(X_1, \dots, X_C)$, the real distribution of our dataset, as possible.
In the typical GAN framework, it may be difficult to find the best optimization solution for such a complex goal, which depends on the number of channels, duration, and distribution of the data.
This is why we use separate generators $G_{i}$ to learn the marginal distribution of each channel, $p(X_i)$, separately, and then use a central discriminator to force preserving the real correlation between the channels by focusing on the conditional distributions $p(X_i|X_{i\neq j})$, where $X_{i\neq j}$ refers to all the channels excluding channel $i$. Figure \ref{fig:COSCI-GAN} depicts this architecture, with more details discussed in the next section.
Essentially, we have two objectives: a local and a central one.

\paragraph{Local objective:} In the local objective, the goal is to estimate the marginal distribution of each channel, $p(X_i)$; which means for each channel $i$, we should optimize:
\begin{equation}
    \min_{q} D(p(X_i) || q(X_i))
\end{equation}
where D is any suitable measure of the distance between two distributions.

\paragraph{Central objective:} In the central objective, the goal is to estimate the conditional distribution of a channel given all the other ones, $p(X_i|X_{i \neq j})$. In order to illustrate this objective, we demonstrate it in a special case where the channels are independent of one another, which means we will have:
\begin{equation}
    \min_q D(\prod_{i=1}^C p(X_i|X_{i \neq j}) || \prod_{i=1}^C q(X_i|X_{i \neq j}))
\label{eqn:centralgeneric}
\end{equation}

Our approach is that each generator $G_i$ share the same initial (noise) source $z$, such that Equation \ref{eqn:centralgeneric} becomes

\begin{equation}
    \min_q D(\prod_{i=1}^C p(X_i|z) || \prod_{i=1}^C q(X_i|z))
\label{eqn:centralCOSCI-GAN}
\end{equation}

We define the global loss to be a linear combination of the loss of local objective and loss of central objective.

\section{COSCI-GAN}
\label{COSCI-GAN}

As shown in Figure \ref{fig:COSCI-GAN}, COSCI-GAN is made up of two main parts: 1) {\em Channel GANs}, which contain pairs of generator-discriminator dedicated to a single channel (univariate TS), and 2) the  {\em Central Discriminator}, dedicated to all channels at once.
Each of these parts is responsible for a specific task.
In channel GANs, the generators are responsible for producing realistic TS and the discriminators are responsible for distinguishing between real and synthetic TS.
The central discriminator is responsible for enforcing that all the generated TS of a given instance have the same correlation as those from real MTS.
Producing all channels simultaneously necessitates that the generative model learn the joint distribution of all TSs, a difficult task that requires a substantial amount of data and time. In contrast, learning the marginal distribution of a single channel is a significantly simpler task. Consequently, the primary purpose of employing channel GANs as opposed to a single multichannel generator-discriminator pair is to assist each individual TS generator in synthesizing more accurate TSs from its own channel’s data distribution. By including the central discriminator, we aim to enforce realistic correlations between the channels as much as possible.
 
 \vspace{-10pt}

\subsection{Algorithm}
\label{COSCI-GAN-algorithm}
 \vspace{-5pt}
Let our multivariate time series have dimensions of $N*L*C$, where $N$ is the number of instances in the dataset, $L$ is the length of each time series, and $C$ is the number of channels. As shown in Figure \ref{fig:COSCI-GAN}, there are $C$ pairs of Generator-Discriminator, or channel GANs. All generators are fed a shared noise vector to begin the generation process. Each generator in a channel GAN will synthesize a TS, and both the generated TS and the corresponding channel of real TS will be passed to their paired discriminator, which determines whether the generated TS is from the same distribution as the real ones. A pseudo-code of the COSCI-GAN algorithm is provided in Algorithm \ref{alg:COSCI-GAN}.
\vspace{-5pt}
\begin{algorithm}[H]
\newcommand{\pluseq}{\mathrel{+}=}
\caption{COSCI-GAN}\label{alg:COSCI-GAN}
\begin{algorithmic}
\For{\texttt{epoch in epochs}}
    \For{\texttt{batch in training set}}
        \State \texttt{Create a noise vector, $Z$.}
        \For{\texttt{i = 1 to $n_{channels}$}}
            \State \texttt{Extract $signal_i$ from the batch.}
            \State \texttt{Generate Fake signals $generated_i$ from $Generator_i$}
            \State \texttt{Train $Discriminator_i$ by feeding $signal_i$ and $generated_i$.}
        \EndFor
        \State \texttt{Train $Central$ $Discriminator$ by feeding }
        \State \texttt{\hspace{20pt}$((generated_1, ..., generated_{n_{channels}}),(signal_1, ..., signal_{n_{channels}}))$.}
        \For{\texttt{i = 1 to $n_{channels}$}}
            \State \texttt{Train $Generator_i$ with $Loss_{D_i}$ and $Loss_{CD}$.}
        \EndFor
    \EndFor
\EndFor
\end{algorithmic}
\end{algorithm}

As mentioned earlier, the central discriminator's role is to preserve the inter-channel correlations. The TS synthesized by all channel generators will be concatenated as an MTS and fed to the central discriminator, which aims to determine whether the MTS is real or fake. We hypothesize that this will penalize unrealistic (un)correlation patterns between channels, and we have provided some evidences in Figure \ref{fig:correlation_1_2} that other State-Of-the-Art (SOTA) methods often exaggerate or even create unrealistic correlations when compared to real data correlation.

\subsection{Training}
\label{COSCI-GAN-training}
During training, the discriminators in the channel GANs (which we will refer to as channel Discriminators from now on), their paired generators, and the Central Discriminator will engage in a three-player game.
The three-player objective of a given channel GAN combined with the central discriminator is:

\begin{equation}
\begin{alignedat}{2}
    \min_{\theta_i} \max_{\phi_i} \max_{\alpha} V(G_{i,\theta_i}, D_{i, \phi_i}, CD_{\alpha}) = &\E_{x \sim P_{data}} &&[log(D_{i, \phi_i}(x_i)])+ \gamma \cdot log(CD_{\alpha}(x))] \\
    & + \E_{z \sim P_z} [&&log(1 - D_{i, \phi_i}(G_{i,\theta_i}(z)) \\
    &  && +\gamma \cdot log(1 - CD_{\alpha}(G_{i,\theta_i}(z), G_{j \neq i}(z))] 
\end{alignedat}
\label{eqn:objectivechannelgan}
\end{equation}

where $G_{i,\theta_i}$ is the $i$-th generator with parameters $\theta_i$, $D_{i, \phi_i}$ is the $i$-th channel discriminator with parameters $\phi_i$, $CD_{\alpha}$ is the central discriminator with the parameters $\alpha$, $P_{data}$ is the distribution of the real time series, $x_i$ is the $i$-th channel of time series $x$, $G_{j \neq i}$ are all the other generators with fix parameters for the optimization step of $G_{i,\theta_i}$,
$\gamma$ is a hyper-parameter that control the trade-off between well-preserving the correlation among the channels versus generating better quality signal within each channel, and $z$ is the shared noise vector sampled from $P_z$ distribution. The objective for all channels is a $2C + 1$ player game adding the terms with subscript $i$ in Equation \ref{eqn:objectivechannelgan}.

In each epoch, we divide the MTS training dataset into a number of batches. If the batch size is assumed to be m, then each batch contains $x^{(1)}, \dots, x^{(m)} \sim D$. For each batch, we sampled as many noise vector $z \sim P_z$ as the batch size.
All channels’ generators synthesize signals, and a gradient ascent step on the discriminators’ parameters $\phi_i$s are taken. Then we concatenate the synthetic signals of all channels and use them in addition to the real data to take a gradient ascent step on the central discriminator parameters $\alpha$. In the end, we take a gradient descent step on the generators parameters $\theta_i$s.

\begin{figure*}[ht!]
    \centering
    \begin{subfigure}[t]{0.68\textwidth}
        \centering
          \includegraphics[width=\linewidth]{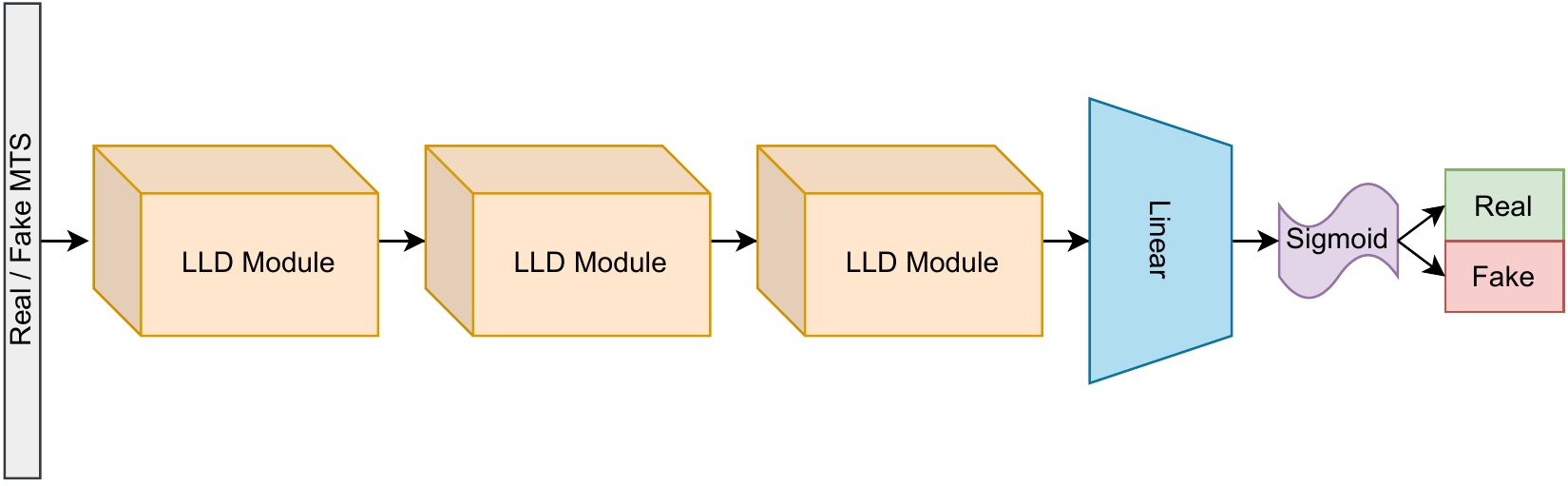}
          \caption{Structure of Central Discriminator.}
          \label{fig:CD}
    \end{subfigure}%
    ~ 
    \begin{subfigure}[t]{0.30\textwidth}
        \centering
        \includegraphics[scale=0.5]{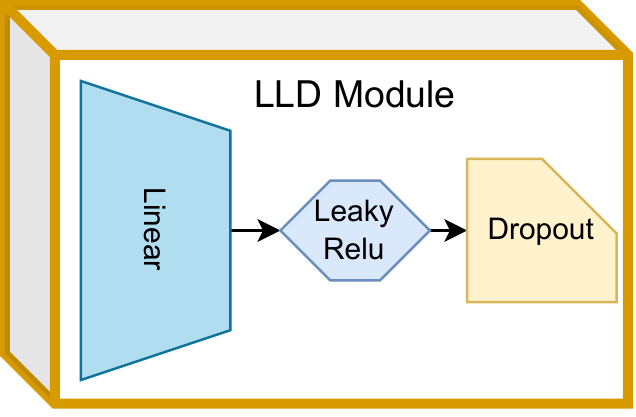}
        \caption{Inside the LLD module.}
        \label{fig:CD_b}
    \end{subfigure}
    \caption{Structure of an MLP-based Central Discriminator.}
    \label{fig:MLP-based_CD}
    \vspace{-6mm}
\end{figure*}

\subsection{Key Implementation Details}
\label{COSCI-GAN-implementation}
There are many possible choices of network types for channel GANs.
We show in Appendix \ref{sec:appen-COSCI-GAN-implementation} that networks based on Long short-term memory (LSTM) generate signals of higher quality than the networks based on Multi-layer perceptron (MLP). 

We have investigated LSTM and MLP based networks for the central discriminator. Although LSTM-based networks perform better than MLP-based networks in Channel GANs, MLP-based networks perform better in the Central Discriminator.
We hypothesize that if the central discriminator is too powerful, the results will be of lower quality, as the generators will strive to make the signals more correlated at the expense of realistic individual TS.

Figure \ref{fig:CD} depicts the structure of an MLP-based central discriminator. It consists of three Linear-LeakyReLU-Dropout (LLD) modules, a linear module, and a sigmoid function. Figure \ref{fig:CD_b} demonstrates the structure of an LLD module.

\section{Empirical Evaluation}
\label{Experiments}
\subsection{Toy Sine Datasets: Diversity vs Correlation Preservation}
\label{sec:Toy-diversity}
To be able to investigate the empirical behavior of COSCI-GAN and, in particular, the effect of the central discriminator, we need to have complete control over the nature of the datasets, particularly the ground truth of our desired tasks. Thus, we simulated three "toy medical" datasets with two channels and used them as "real" datasets to generate synthetic data. For all these datasets, we assumed that each instance correspond to a different patient, and each patient produce measurements for two channels ($c_1$ and $c_2$). To make the datasets a bit more realistic, we also assumed that there are two types of patients ($pt_1$ and $pt_2$), as in "healthy" vs "condition". The three datasets are:

\begin{itemize}
    \item \textbf{Simple Sine} is derived from the formula: $x = A \sin(2\pi ft)+ \epsilon$. The difference between the signals is that the amplitude ($A$) for patient type 1 comes from $ \mathcal{N}(0.4,\,0.05)\,$, whereas the amplitude for patient type 2 comes from $ \mathcal{N}(0.6,\,0.05)\,$. The other difference is that Channel 1 has a frequency ($f$) of $0.01$, while channel 2 has a frequency of $0.005$. In addition, the noise $\epsilon$ comes from $ \mathcal{N}(0,\,0.05)\,$.
    
    \item \textbf{Sine with frequency changes} contains the same signals as \textit{Simple Sine}, except that the frequency of all sine functions doubles exactly in the middle of the time series. This allows us to examine the situation with varying frequencies. 
    
    \item \textbf{Anomalies} is created by replacing the middle of the time series with Gaussian noise, thus allowing us to examine the impact of anomalies.
\end{itemize}
The visualization of all three toy datasets are 
available in Appendix \ref{sec:appen-Toy-vis}.

We assessed the behaviour of COSCI-GAN, particularly the central discriminator,  by two criteria:

(1) Diversity: requiring that the generators should synthesize from both patients’ distributions and there should be no {\em mode collapse}, a common failure of GANs, which occurs when the generator fails to produce results as diverse as the real data. We measured the diversity by comparing the distribution of amplitude of patients’ signals in the real dataset, which is a bimodal Gaussian distribution, with the distributions of amplitude of the generated samples using Wasserstein Distance (WD). We took the WDs average (AWD) to aggregate across the channels. A lower AWD indicates a closer similarity to the real distributions. The first column of Table \ref{table:wd-table} shows the AWD for the various cases.

(2) Correlation Preservation: requiring that the amplitudes of channels for each patient types should be equal to each other as much as possible. We measured the amplitudes of channel 1 and 2 in all signals and verified their similarity. We defined our correlation metric as the average euclidean distance (AED) between the amplitude mapped on a 2D plane (Channel 1 vs Channel 2) and a line with slope 1. The resulting plot is provided in Appendix \ref{sec:appen-Toy-diversity}, and the numeric values are summarized in Table \ref{table:wd-table}. A lower AED indicates stronger preservation of correlation. 

\begin{table}[htp]
    \vspace{-3mm}
  \caption{Results of Diversity (AWD) V.S. Correlation Analysis (AED)}
  \label{table:wd-table}
  \centering
  \begin{tabular}{c|ccccc}
    \toprule

    Dataset & Method & AWD & AED  \\
    \midrule

    Simple Sine & Without CD & \textbf{0.0472} & 0.1326  \\
     & With CD & 0.0800 & \textbf{0.0177} \\
    Freq changes & Without CD & \textbf{0.0397} & 0.0769 \\
     & With CD & 0.0679  & \textbf{0.0242} \\
    Anomalies & Without CD  & \textbf{0.0540} & 0.0766 \\
     & With CD  & 0.0726 & \textbf{0.0161} \\
    \bottomrule
  \end{tabular}
  \vspace{-2mm}
\end{table}

As clearly shown in Table \ref{table:wd-table}, there is a trade-off between diversity and correlation preservation. That is when we used the CD, the correlation was better preserved, but at the expense of diversity (similarity between the generated time series’ marginal distribution of amplitudes and the toy Simple Sine time series’ marginal distribution of amplitudes). Conversely, not using the CD would allow the generated sample distributions to be closer to the real data, but the generated channels were less correlated. The strength of the CD is controlled by the parameter $\gamma$ as shown in Equation (\ref{eqn:objectivechannelgan}). As we conducted experiments to tune the hyper-parameter $\gamma$, we observed that there is a stable range for $\gamma$. We ended up setting $\gamma$ to $5$, which provides stable results for all the experiments presented in this paper, whether the datasets are the toy medical ones or the real ones to be discussed later.

\vspace{-0.2cm}

\subsection{Toy Sine Datasets - Feature-based Correlation Analysis}
\label{sec:correlation-analysis}

\vspace{-0.2cm}

The {\em catch22} feature set has been introduced to capture 22 CAnonical Time series CHaracteristics commonly seen in diverse time series data mining tasks \cite{Lubba2019catch22CT}. Using this feature set, we assessed how correlation between any pair of catch 22 features was preserved in synthetic time series data generation. In other words, if a pair is strongly correlated in the real dataset between the two channels, we would like to see that preserved in the synthetic dataset. Similarly, if two features are not correlated in the real dataset, they should remain uncorrelated in the synthetic dataset. 

\begin{figure}[htp]
  \centering
  \includegraphics[width=\linewidth]{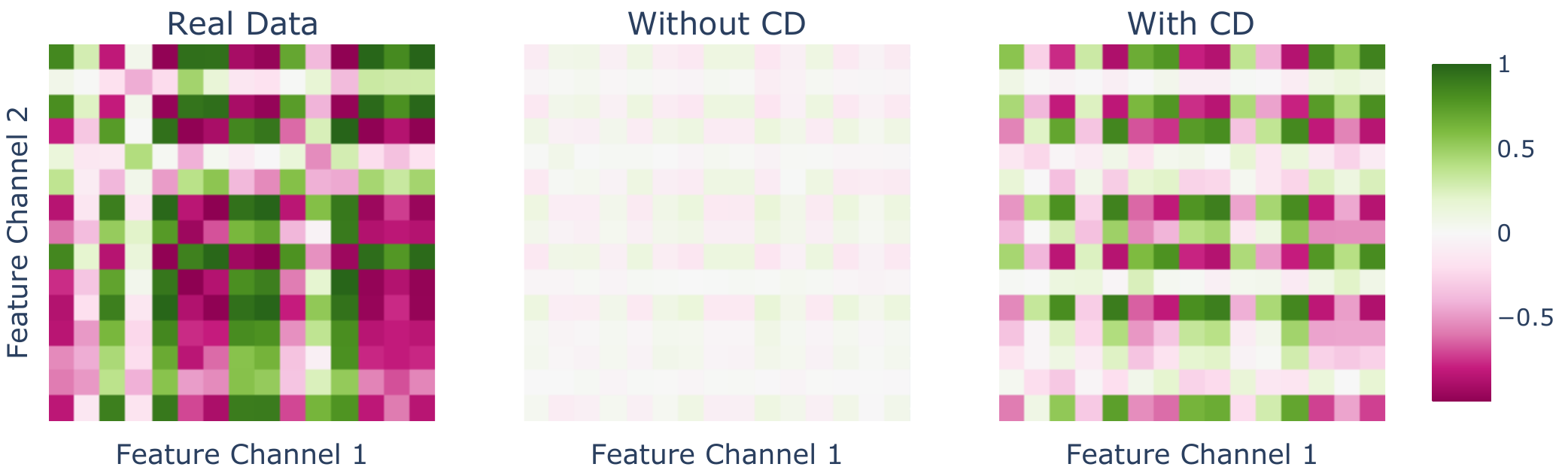}
  \caption{Heatmaps for Catch22 Pairwise Correlations.}
  \label{fig:correlation}
\end{figure}

Figure \ref{fig:correlation} shows three heatmaps for those pairwise correlations between the two channels of simple sine dataset (The same figure for other toy datasets are provided in Appendix \ref{sec:appen-Toy-correlation}). To simplify the heatmaps, we removed 7 features that were constant among the real datasets, and kept the remaining 15 features that varied.
The left heatmap shows the pairwise correlations of the 15 features for the real dataset. The centre heatmap is the one for the synthetic dataset generated by COSCI-GAN without a CD, whereas the right heatmap is the one generated by COSCI-GAN having the CD. Clearly, the right heatmap resembles the left heatmap much more closely. The centre heatmap shows that without the CD, almost all the correlation relationships of the 15 features were destroyed. 

While the heatmaps are useful for visualization, we also compared quantitatively the correlation matrices between the two channels using various metrics: (1) Mean Absolute Error (MAE), (2) Frobenius norm, (3) Spearman's $\rho$, and (4) Kendall's $\tau$. For MAE and the Frobenius norm, a smaller value indicates greater similarity between the correlation matrices of the real and synthetic datasets.
For Spearman's coefficient and Kendall's coefficient, the closer the value is to 1, the higher is the similarity.
Results shown in Table \ref{table:correlation-table} provide convincing evidence of the effectiveness of the CD in synthetic MTS generation.

\begin{table}[htp]
    \vspace{-6mm}
  \caption{Similarity between Correlation Matrices}
  \label{table:correlation-table}
  \centering
  \begin{tabular}{c|ccccc}
    \toprule

    Dataset & Method & MAE & Frobenius norm & Spearman’s $\rho$ & Kendall’s $\tau$ \\
    \midrule

    Simple Sine & Without CD & 0.671 & 11.295 & -0.761 & -0.569 \\
     & With CD  & \textbf{0.298} & \textbf{5.666} & \textbf{0.848} & \textbf{0.700}  \\
    Freq changes & Without CD & 0.268 & 7.889 & 0.259 & 0.174    \\
     & With CD  & \textbf{0.131} & \textbf{3.413} & \textbf{0.834} & \textbf{0.661} \\
    Anomalies     & Without CD & 0.289 & 8.113 & 0.428 & 0.297\\
     & With CD  & \textbf{0.199} & \textbf{5.362} & \textbf{0.786} & \textbf{0.612} \\
    \bottomrule
  \end{tabular}
  \vspace{-4mm}
\end{table}

\subsection{EEG Eye State Dataset and Downstream Classification}
\label{sec:EEG}
We selected a 14-channel EEG eye state dataset to measure the effectiveness of COSCI-GAN on real signals \cite{Dua:2019}\footnote{\href{https://archive.ics.uci.edu/ml/datasets/EEG+Eye+State}{https://archive.ics.uci.edu/ml/datasets/EEG+Eye+State} with license details therein}.
This dataset contains a label indicating whether the patient's eyes were open or closed (1 indicates closed, and 0 indicates open).
Each time series is 117 seconds long, resulting in 14980 samples at a sampling rate of 128 per second. To remove outliers from the dataset, we experimented with various z-score values and determined that a value of 3 is optimal for our dataset, so we removed points with z-scores greater than 3, the same value as other EEG studies (e.g. \cite{8919795}).
The label of the dataset allows us to create a downstream eye blink classification task as follows. 

We extracted a window of 800 samples in length containing an eye blink, with a margin of 200 samples at the beginning and end of each eye blink, and labelled those frames with 1.
We also extracted 800 samples that did not contain an eye blink, with a margin of 200 samples between the beginning and end of an eye blink, and labelled them with 0.
We now have 1024 frames of each label for classification.
Then we performed a forward feature selection, and chose top 5 channels regarding the accuracy of our classification task.

To measure the effectiveness of COSCI-GAN for classification, we used the approach of train-on-real and test-on-fake, and the opposite approach of train-on-fake and test-on-real. We measured the accuracy of an LSTM-based classifier, which is described in details in Appendix \ref{sec:appen-EEG-classifier}, on a dataset that contains two channels.
Table \ref{table:classification-1-table} shows the results of this comparison. We compared the accuracy of the classifier when the synthetic data were generated with and without the CD. Once again, the CD brings significant value to downstream classification tasks.

\begin{table}[htp]
    \vspace{-4mm}
  \caption{Accuracy in Classification Task}
  \label{table:classification-1-table}
  \centering
  \begin{tabular}{ccc}
    \toprule
     Experiment & COSCI-GAN with CD & COSCI-GAN without CD \\
    \midrule
    
    Train-on-real, Test-on-fake & 0.790 & 0.644 \\
    Train-on-fake, Test-on-real & 0.634 & 0.561 \\
    \bottomrule
  \end{tabular}
  \vspace{-2mm}
\end{table}

Next we compared COSCI-GAN against a baseline method for generating MTS data. The baseline method is an LSTM-based GAN that simultaneously generated all channels. The same LSTM network was utilized in the baseline method and the COSCI-GAN modules. The only difference is that the output layer of networks in the baseline method must generate all channels rather than just one. Below we show the results of two experiments: (1) the All-synthetic experiment, and (2) the Augmentation experiment. 

\begin{figure}[htp]\centering
\subfloat[Accuracy of All-Synthetic Experiment.]
{\label{fig:All_fake}\includegraphics[width=.425\linewidth]{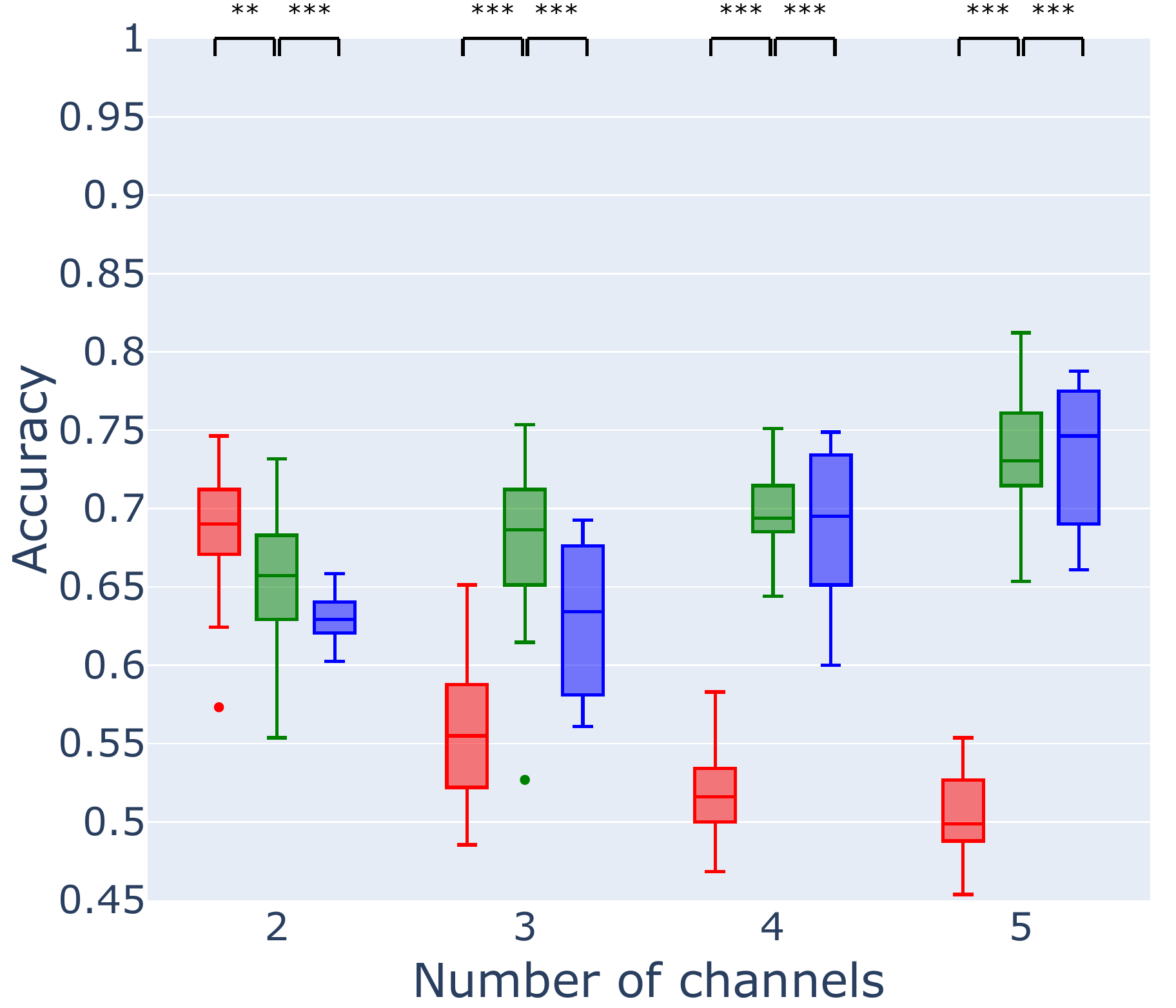}}\hfill
\subfloat[Accuracy of Augmentation Experiment.]
{\label{fig:Augmentation}\includegraphics[width=.5\linewidth]{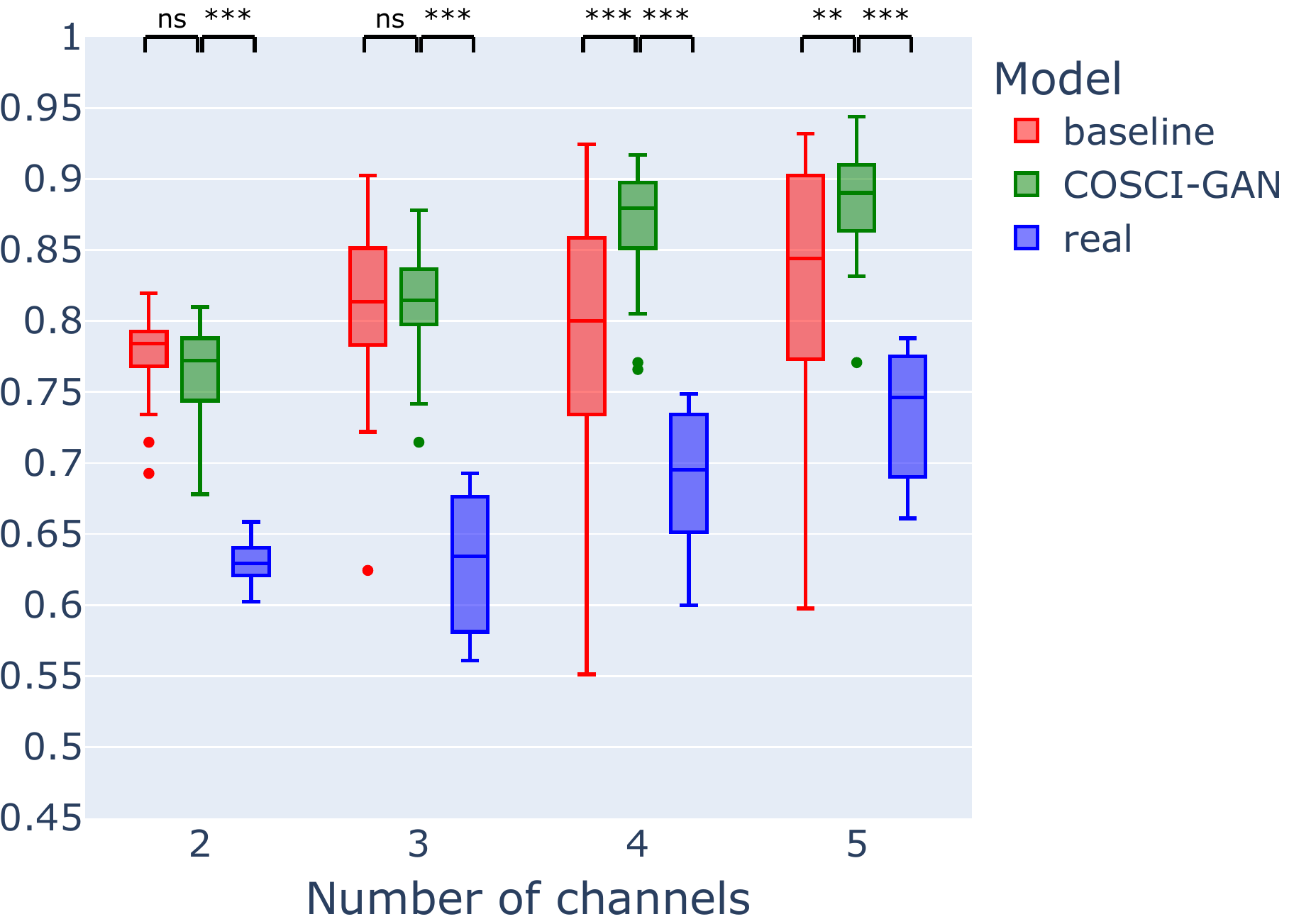}}\par 
\subfloat[Accuracy of Augmentation Experiment: each point of the same color corresponds to various augmentation ratios, from $1:1$ to $1:10$, see text for details.]{\label{fig:scatter}\includegraphics[width=.5\linewidth]{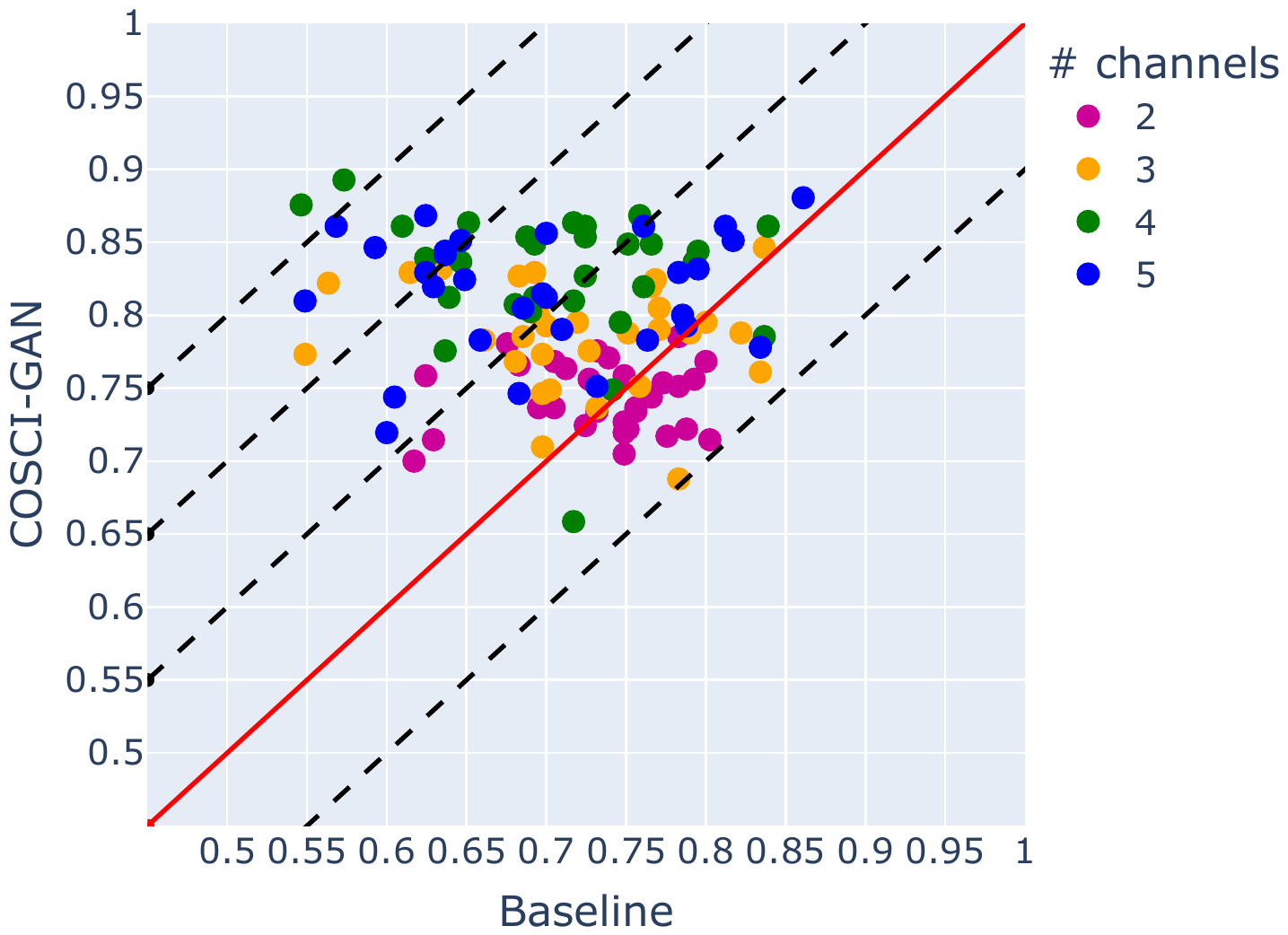}}
\caption{Experiment three results. In figure (a) and (b), T-tests were applied to the results in order to show whether there is a statistically significant difference between the distribution of the results. \textbf{ns} means no significant difference, \textbf{**} means $0.001<$ p-value $<0.01$, and \textbf{***} means p-value $ < 0.001$.}
\label{fig:experiments-all-fake-augmentation}
\vspace{-10pt}
\end{figure}

\paragraph{(1) All-synthetic experiment} In this experiment, we assessed how well COSCI-GAN performed in classification task when compared with the baseline method and the actual dataset. We performed cross-validation by using 80\% of our real dataset for training the GANs. Then only the synthetic data were used to train the classifiers, which were then tested on the hold-out 20\% of the real dataset. We investigate the utility of the synthetic data for different number of channels in Figure \ref{fig:All_fake}. We repeated each experiment 30 times with different random seeds for each setting, and statistical significance tests were done on the boxplots. A further comparison including COSCI-GAN without the CD is  provided in Appendix \ref{sec:appen-EEG-Experiments}.

Figure \ref{fig:All_fake} demonstrates that as the number of channels is increased, the accuracy of a classifier trained on data generated by COSCI-GAN and evaluated on real datasets increases. In contrast, the baseline method went the opposite way and demonstrated that the performance of MTS generation with all channels together degrades as the number of channels increases. As the number of channels grew, synthetic data generated by COSCI-GAN gave a similar average performance as the real dataset.

\paragraph{(2) Augmentation experiment} \label{sec:augmentation} This experiment was set up exactly like the previous one, but instead of using only synthetic time series to train the classifiers, we augmented the real dataset with an equal number of  synthetic training samples.   In Figure \ref{fig:Augmentation}, the boxplots for the real datasets across the different number of channels are exactly the same as those in Figure \ref{fig:All_fake} for the real datasets.

Between Figure \ref{fig:All_fake} and Figure \ref{fig:Augmentation}, COSCI-GAN shows significant improvements in accuracy. The difference was that the synthetic data generated were added to the real data.
COSCI-GAN still outperformed the baseline method both in terms of the median and the variations in accuracy. 

Figure \ref{fig:Augmentation} was based on an augmentation ratio of 1:1. 
Figure \ref{fig:scatter} shows a scatter plot of comparing the accuracy of COSCI-GAN (y-axis) against the baseline method (x-axis) across six different augmentation ratios: $1:1$, $1:2$, $1:4$, $1:6$, $1:8$, $1:10$, i.e. augmenting with up to 10 times more synthetic data than real training data.
We repeated the experiment five times with different random seeds for each of these settings; the accuracy of each run is plotted in Figure \ref{fig:scatter}. Thus, any point above the diagonal red line indicates that COSCI-GAN outperformed the baseline method. The points in the figure are colour-coded based on the number of channels. It is obvious that COSCI-GAN dominated the baseline method when there is more than two channels. To further quantify the differences in accuracy, we add dotted diagonal lines in \ref{fig:scatter}, representing a difference in accuracy with increments of 0.1. For instance, the lowest dotted diagonal line represents the cases when the accuracy of COSCI-GAN is 0.1 below that of the baseline method. Conversely, the other three dotted diagonal lines represent the situations when the accuracy of COSCI-GAN is 0.1, 0.2 or 0.3 better than that of the baseline. Figure \ref{fig:scatter} shows that COSCI-GAN almost always performed better than the baseline method, up to 0.3 higher accuracy, and most exceptions are from the 2 channel dataset where the accuracy is at most 0.1 lower.

\vspace{-0.3cm}

\subsection{Comparing with State-of-the-art methods on EEG Classification}
\label{sec:sota}
\vspace{-0.2cm}

In the final experiment, we compared COSCI-GAN with two state-of-the-art (SOTA) methods: TimeGAN\footnote{\href{https://github.com/jsyoon0823/TimeGAN}{https://github.com/jsyoon0823/TimeGAN} with license details therein} \cite{Yoon2019TimeseriesGA} and the most recent  Fourier Flows \footnote{\href{https://github.com/ahmedmalaa/Fourier-flows}{https://github.com/ahmedmalaa/Fourier-flows} with license details therein} \cite{Alaa2021GenerativeTM} discussed previously.
In these papers, the downstream task was predicting the next time point of the time series; i.e., forecasting.
Because our focus is on classification, we used TimeGAN and Fourier Flows' code to generate the five EEG channels that we chose in the previous experiment and performed the augmentation experiment that we described in \ref{sec:augmentation}. 
As shown in Figure \ref{fig:augmentation_sota} COSCI-GAN leads to the best classification accuracy and there is a statistically significant difference between the results of COSCI-GAN and two other methods.

To compare the statistical properties of COSCI-GAN with SOTA methods, we repeated the correlation analysis from \ref{sec:correlation-analysis}.
Figures similar to Figure \ref{fig:correlation} for TimeGAN and Fourier Flows are provided in Appendix \ref{sec:appen-SOTA-Correlation}.
We computed the MAE between the real correlation matrix and each method's correlation matrix for each pair of channels in the EEG dataset.
Table \ref{table:correlation-sota} shows that COSCI-GAN, in addition to giving better classification accuracy, provides the closest similarity in inter-channel correlations to the real dataset. 
  
\begin{table*}[h]
\vspace{-10pt}
    \begin{minipage}{0.59\linewidth}
        \begin{figure}[H]
            \centering
            \includegraphics[width=\linewidth]{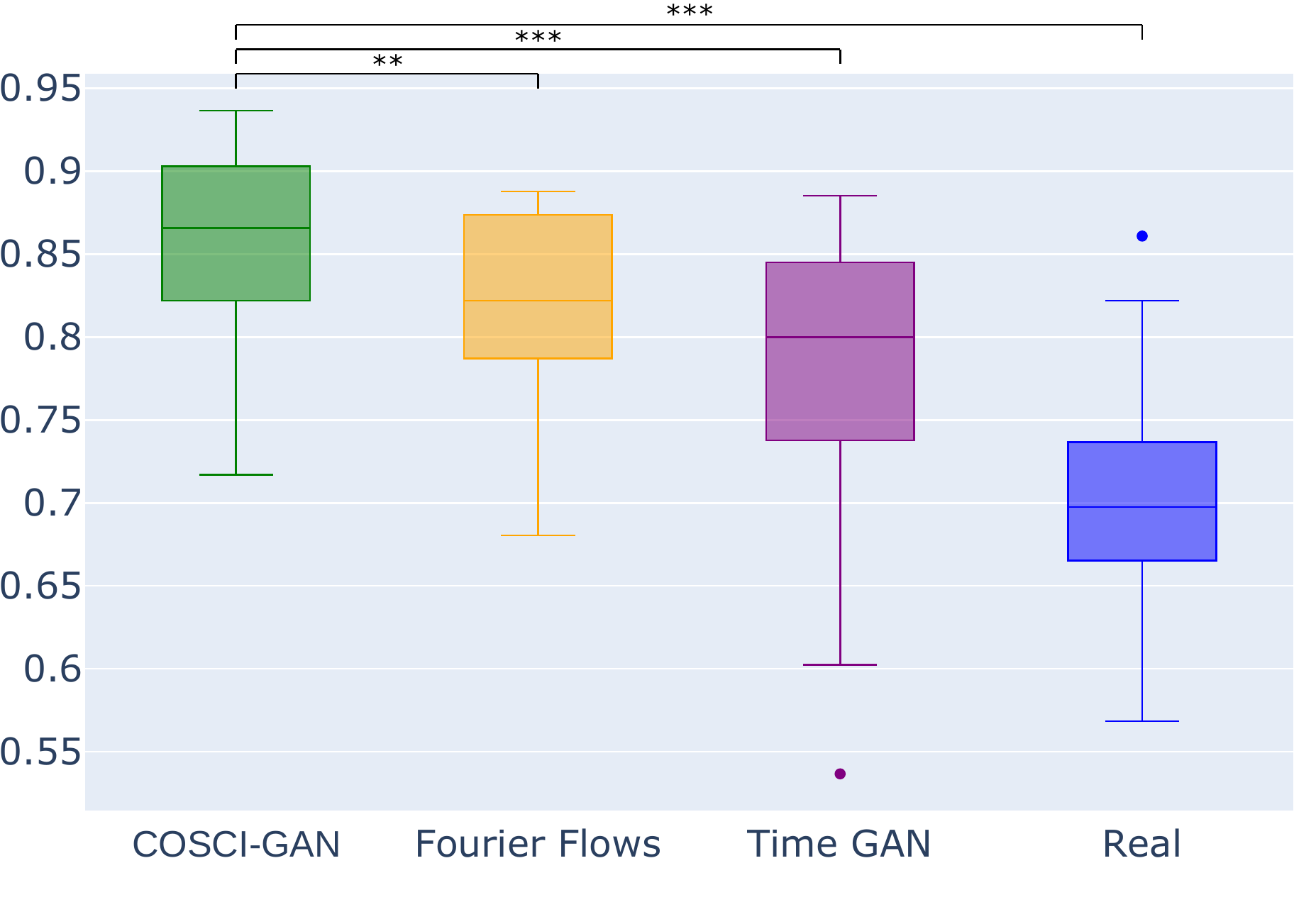}
            \caption{\centering Accuracy for the Augmentation Experiment Comparison with SOTA methods}
            \label{fig:augmentation_sota}
        \end{figure}
    \end{minipage}
    \hfill
    \begin{minipage}{0.38\linewidth}
        \caption{\centering Similarity between Correlation Matrices in EEG dataset mean $\pm$ standard deviation}
        \label{table:correlation-sota}
        \centering
        
        \begin{tabular}{cc}
            \toprule
            Method & MAE \\
            \midrule
            COSCI-GAN & \textbf{0.111} $\pm$ \textbf{0.005}\\
            TimeGAN & 0.257 $\pm$ 0.008\\
            Fourier Flows & 0.146 $\pm$ 0.006\\
            \bottomrule
        \end{tabular}
    \end{minipage}    
    \vspace{-2pt}
\end{table*}
In TimeGAN and Fourier Flows papers, a daily historical Google stocks dataset from 2004 to 2019 was used, and PCA\cite{Bryant1995PrincipalcomponentsAA} and t-SNE\cite{vanDerMaaten2008} plots were shown to compare their diversity in compare with the real data. We repeated this experiment using data from TimeGAN repository \footnote{\href{https://github.com/jsyoon0823/TimeGAN/blob/master/data/stock_data.csv}{https://github.com/jsyoon0823/TimeGAN/blob/master/data/stock\_data.csv}}, and showed that COSCI-GAN samples were more diverse and distributed in a way that was more similar to the real dataset distribution. The figures of this experiment are provided in Appendix \ref{sec:appen-SOTA-PCA}.

\section{Discussion}
\label{Discussion}

In this paper, we introduced COSCI-GAN, a novel framework for multivariate time-series generation that delivers more correlated channels.
By preserving the correlation between channels, COSCI-GAN is able to generate time-series that are more similar to the real time-series and achieve better performance in downstream classification tasks than other state of the art methods.

We have shown that our framework is relevant for generating MTS from a common source and we argue that it is particularly suited for human-based biometric measurements.
In our experiments, we have never had performance limitations, we foresee however, that COSCI-GAN will not scale to a very large number of channels as a dedicated GAN for each channel is needed. However, this issue is not exclusive to the COSCI-GAN method. It is a "computing resources" limitation for the COSCI-GAN method, whereas it is a fundamentally intractable problem for the baseline and many other methods. In addition, COSCI-GAN has the benefit of being parallelizable, which makes it faster.
On the other extreme, we have shown that COSCI-GAN is not competitive for two channels where it often performed worse than a simple baseline.

It is worth re-iterating that synthetic data generation does \textit{not} guarantee privacy and similarly, there is no way to know in advance the performance on a downstream task so both characteristics should be empirically evaluated post-generation.
Outliers and minorities are often affected most by privacy leaks as they do not get protected by a large number of similar data samples.
We also acknowledge that synthetic data generation can cause harm propagating and even magnifying bias from the data it is based on. 

As future work, our method could be extended to more practical use cases where the various channels corresponds to different types of time series, e.g. heartbeats, temperature, respiration and wearable measurements and so on.
As another extension, we could consider having an initial noise embedding (corresponding to several initial noises) that are all originated from a single source, in order to have more control on each channel's distribution.
On the technical side, our framework can be implemented with a wide variety of GANs chosen based on the data type including modern architectures such as transformers. Additionally, as stated previously, Channel GANs could train in parallel, which would accelerate the training process. The code we have provided is modular and its parallelized version can be implemented in future.


\medskip
\bibliographystyle{unsrtnat}

{
\small
\bibliography{citations}

}

\newpage

\appendix
\begin{appendix}
\section{COSCI-GAN}

\subsection{Training}
\label{sec:appen-COSCI-GAN-training}

All codes and experiments are available in the COSCI-GAN repository\footnote{\url{https://github.com/aliseyfi75/COSCI-GAN}}. Before beginning the training, several hyper-parameters can be adjusted:

\begin{itemize}[noitemsep]
    \item \textbf{criterion} determines the criterion in training and can be either Binary Cross Entropy (BCE) or Mean Squared Error (MSE).
    \item \textbf{CD\_type} will determine the type of central discriminator, which will be one of Multi-Layer Perceptron (MLP) or Long Short-Term Memory (LSTM), as discussed in Section \ref{COSCI-GAN-implementation}.
    \item \textbf{LSTMG} and \textbf{LSTMD} are two flags that determine the type of Channel Generators and Discriminators. If those two flags set to \textbf{True}, an LSTM-based network will be used for the Channel generators and Discriminators; Otherwise, and MLP-based netweork will be used.
    \item \textbf{withCD} flag controls whether a Central Discriminator is used in the COSCI-GAN structure.
    \item \textbf{nepochs} is the number of epochs and \textbf{batch\_size} is the batch\_size in the training procedure as described in Algorithm \ref{alg:COSCI-GAN}. 
    \item \textbf{glr}, \textbf{dlr}, and \textbf{cdlr} are the learning rate of the channel generators, discriminators, and central discriminator respectively.
    \item \textbf{real\_data\_fraction} is the fraction of the real dataset that is used to train the COSCI-GAN.
    \item \textbf{gamma}, $\gamma$, is the tuning parameter that controls the trade-off between well-preserving the correlation between the channels and generating higher-quality signal within each channel, as described in the Section \ref{COSCI-GAN-training}.
    \item \textbf{noise\_len} is the length of noise vector that channel GANs share with each other.
    \item \textbf{nsamples} is the length of the multi-variate time-series in the dataset.
    \item \textbf{Ngroups} is the desired number of channels to generate.
    
\end{itemize}

The Table \ref{table:hyper-parameters} shows the default values for the aforementioned hyper-parameters. These values were chosen based on our experiments and heuristics.

\begin{table}[htp]
\vspace{-4mm}
  \caption{Default values of hyper-parameters}
  \label{table:hyper-parameters}
  \centering
  \begin{tabular}{c|ccccc}
    \toprule

    Hyper-Parameter & criterion & CD\_type & LSTMG & LSTMD & withCD  \\
    \midrule

    Value & BCE & MLP & True & True & True \\
    \specialrule{.15em}{.15em}{.15em}
    Hyper-Parameter & nepochs & batch\_size & glr & dlr & cdlr \\
    \midrule

    Value & $100$ & $32$ & $10^{-3}$ & $10^{-3}$ & $10^{-4}$  \\
    
    \specialrule{.15em}{.15em}{.15em}
 
    Hyper-Parameter & real\_data\_fraction & gamma & noise\_len & nsamples & Ngroups  \\
    \midrule

    Value & 1.0 & 5.0 & 32 & \multicolumn{2}{c}{\hspace{-7pt} Depends on the Real Dataset} \\
    
    \bottomrule
  \end{tabular}
\end{table}

COSCI-GAN should receive a 2D array as the real dataset, in the following format: If the real multi-variate time-series have a dimension of $N \times L \times C$, as described in Section \ref{COSCI-GAN-algorithm}, then the dataset's channels should be concatenated with each other to form a 2D matrix with shape $N \times C*L$ before using the COSCI-GAN. During the training process, each channel will be extracted from this 2D matrix.

We ran our experiments on four NVIDIA P100 Pascal GPU cores and four Intel E5-2650 v4 Broadwell @ 2.2GHz CPU cores. COSCI-GAN's execution time is highly dependent on the number of instances in the dataset, the length of the time-series, the number of channels, and the hyper-parameters. Table \ref{table:running-time} shows the COSCI-GAN running time for various datasets using the default hyper-parameter values from Table \ref{table:hyper-parameters}. $\#$Samples is the number of samples in the time-series, or the length of the time-series; And m:s stands for minutes:seconds.

\begin{table}[htp]
    \vspace{-4mm}
  \caption{Running Time of COSCI-GAN}
  \label{table:running-time}
  \centering
  \begin{tabular}{c|cccc}
    \toprule

    Dataset & $\#$Instances & $\#$Samples & $\#$ Channels & Running Time (m:s)\\
    \midrule
    
    Simple Sine &  2048 & 800 & 2 & 6:53\\
    Sine with Frequency Change & 2048 & 800 & 2 & 6:58\\
    Sine with Anomaly & 2048 & 800 & 2 &  6:59\\
    EEG Eye State & 2048 & 100 & 5 & 10:05\\

    \bottomrule
  \end{tabular}
  \vspace{-4mm}
\end{table}

\subsection{Key Implementation Details}
\label{sec:appen-COSCI-GAN-implementation}

\begin{figure*}[htp]
\begin{subfigure}{0.49\linewidth}
  \includegraphics[scale=0.6]{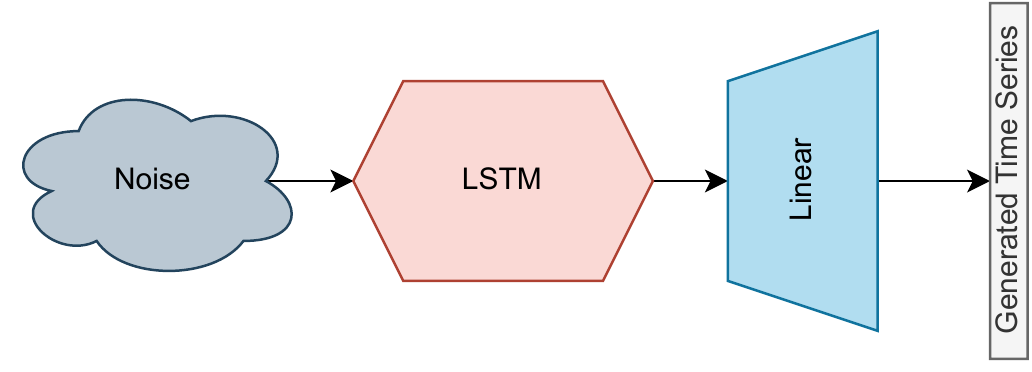}
  \caption{Structure of LSTM-based Generator}
\end{subfigure}
\begin{subfigure}{0.49\linewidth}
  \includegraphics[scale=0.6]{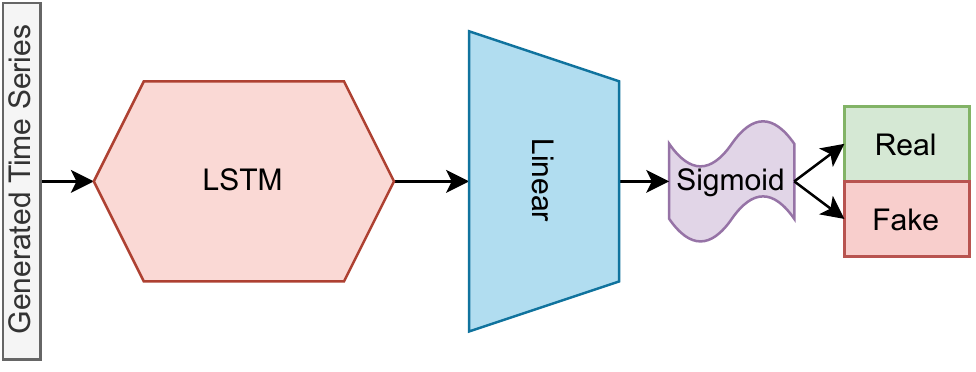}
  \caption{Structure of LSTM-based Discriminator}
\end{subfigure}
\caption{Structure of Channel GAN}
\label{fig:Structure}
\end{figure*}

Figure \ref{fig:Structure} depicts the structure of an LSTM-based Generator and an LSTM-based Discriminator, which together form a Channel GAN. \textbf{Pytorch\cite{NEURIPS2019_9015}} library is used to implement all of the networks in COSCI-GAN. The hidden dimension of the LSTM module in an LSTM-based Generator is $256$, and the number of layers is $1$. Additionally, the linear module maps the output of the LSTM module from its output dimension ($256$) to the desired length of the time series. An identical LSTM module is used in the LSTM-based Discriminator, followed by a linear module that maps the output from the LSTM module's output dimension ($256$) to dimension $1$, a single number, and feeds it to a Sigmoid function.

We have three Linear-LeakyReLU-Dropout (LLD) modules after each other in an MLP-based discriminator, which we use for the Central Discriminator, as shown in Figure \ref{fig:MLP-based_CD}, followed by a linear module and a Sigmoid function at the end. 

The \textbf{negative\_slope} is set to $0.1$ in all Leaky-ReLU functions, and the $p$ in the dropout module is set to $0.3$. The first Linear layer in the LLD module will map the input time series from its length to $256$, the next Linear layer from $256$ to $128$, the next from $128$ to $64$, and the final Linear layer from $64$ to $1$, a single number.

As stated in Section \ref{COSCI-GAN-implementation}, the results of LSTM-based COSCI-GAN outperform MLP-based COSCI-GAN. To compare these two structures, we trained two COSCI-GANs with the same configuration. The only difference was the type of Channel GANs. Then, using the same noise vector, we generate their multivariate time-series. Finally, we computed the correlation matrices described in Section \ref{sec:correlation-analysis} for the generated samples from each of these structures, with and without a Central Discriminator, and summarised the results in Table \ref{table:correlation-table-2}. The results demonstrate the benefit of an LSTM-based network as well as the importance of having a Central Discriminator.

\begin{table}[htp]
  \caption{Similarity between Correlation Matrices - Simple Sine Dataset}
  \label{table:correlation-table-2}
  \centering
  \begin{tabular}{c|ccccc}
    \toprule

    Structure & Method & MAE & Frobenius norm & Spearman’s $\rho$ & Kendall’s $\tau$ \\
    \midrule
    
    \multirow{2}{*}{LSTM} &  With CD & \textbf{0.298} & \textbf{5.666} & \textbf{0.848} & \textbf{0.700}  \\
     & Without CD & 0.671 & 11.295 & -0.761 & -0.569 \\
    
    \multirow{2}{*}{MLP} & With CD & 0.475 & 8.386 & 0.584 & 0.534 \\
     & Without CD & 0.843 & 14.742  & -0.826 &  -0.687  \\
    
    \bottomrule
  \end{tabular}
\end{table}

\section{Empirical Evaluation}

\subsection{Toy Sine Datasets: Diversity vs Correlation Preservation}

\subsubsection{Data Visualization}
\label{sec:appen-Toy-vis}

\begin{figure}[H]\centering
\subfloat[Simple Sine.]
{\label{fig:Simple_Sine}\includegraphics[width=\linewidth]{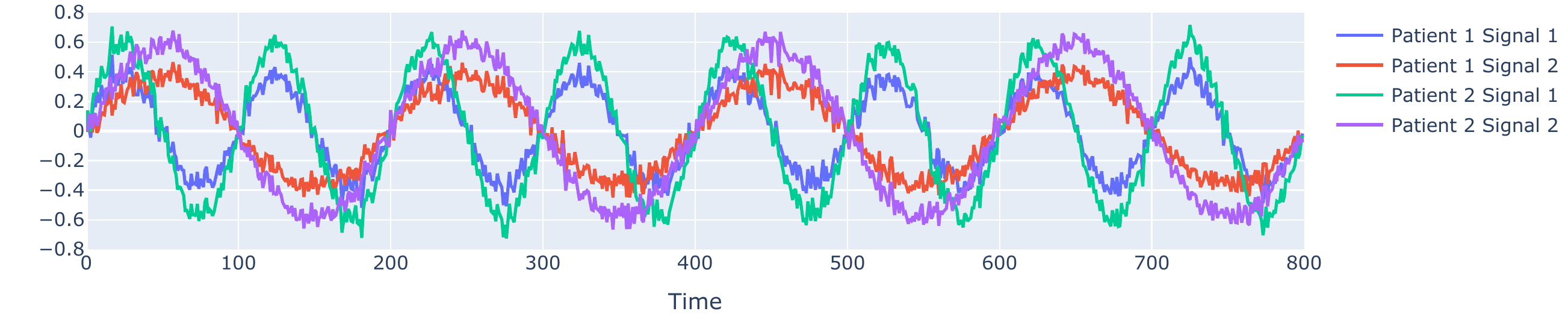}}\par
\subfloat[Sine with Frequency Changes.]
{\label{fig:Sine_with_freq_change}\includegraphics[width=\linewidth]{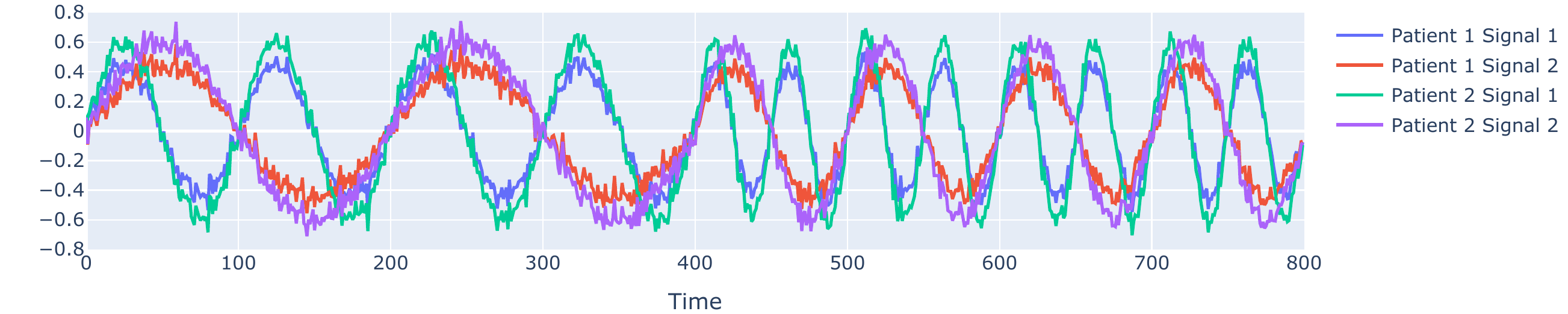}}\par 
\subfloat[Sine with Anomaly.]{\label{fig:Sine_with_anomaly}\includegraphics[width=\linewidth]{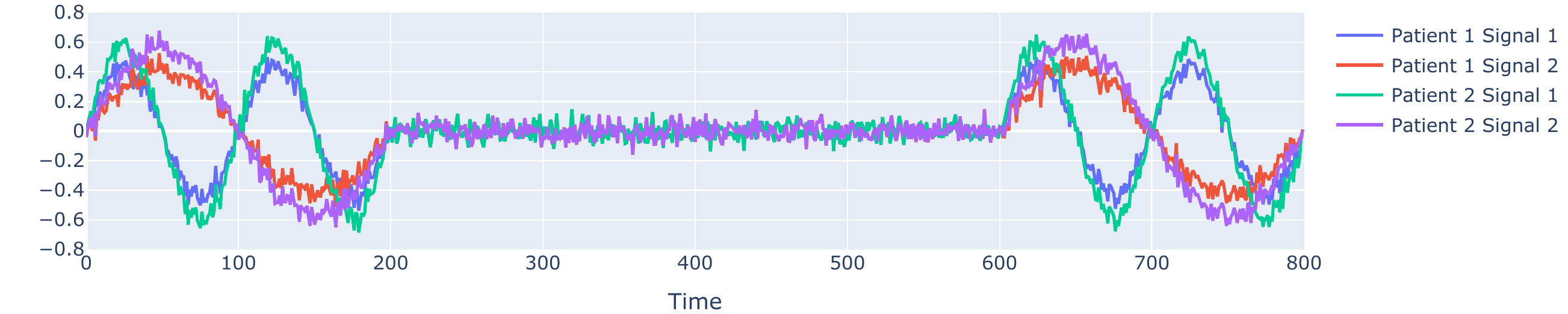}}
\caption{Visualization of Toy Datasets}
\label{fig:visualization}
\end{figure}

Figure \ref{fig:visualization} depicts a visualisation of the toy datasets. As described in Section \ref{sec:Toy-diversity}, each toy dataset contains four signals, two patients that each has a two-channel time-series.

\subsubsection{Diversity V.S. Correlation Analysis}
\label{sec:appen-Toy-diversity}

\begin{figure}[H]
  \centering
  \includegraphics[width=0.4\linewidth]{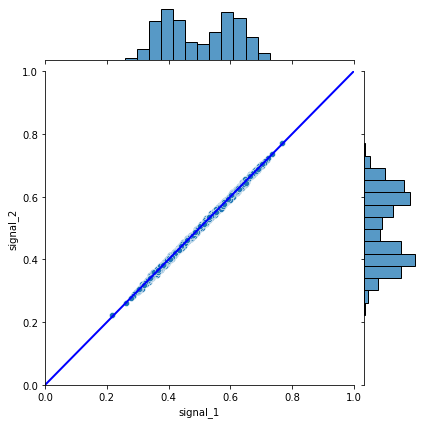}
  \caption{\centering Amplitudes from Simple Sine Toy Dataset. Each point represents the signal amplitude in the first channel (x-axis) vs. the signal amplitude in the second channel (y-axis). The marginal distributions of the amplitudes in the corresponding channel are shown on the side of each axis.}
  \label{fig:wd-real}
\end{figure}

As we demonstrated in Section \ref{sec:Toy-diversity}, there is a trade-off between diversity of generated samples and correlation between generated channels. We went over the Diversity and Correlation Preservation criteria in depth and discussed how we measure each of them in Section \ref{sec:Toy-diversity}. Figure \ref{fig:wd-real} depicts the settings that we described in Section \ref{sec:Toy-diversity} for the simple sine toy dataset . The amplitude of each channel is a bimodal Gaussian distribution, as shown in the figure, and the amplitudes of signals in two channels are nearly identical (The reason that they are not exactly same is that we added noise to each time point of our time-series).

\begin{figure*}[htp]
With CD
\hspace{12pt}
\begin{subfigure}{0.29\linewidth}
  \includegraphics[width=\linewidth]{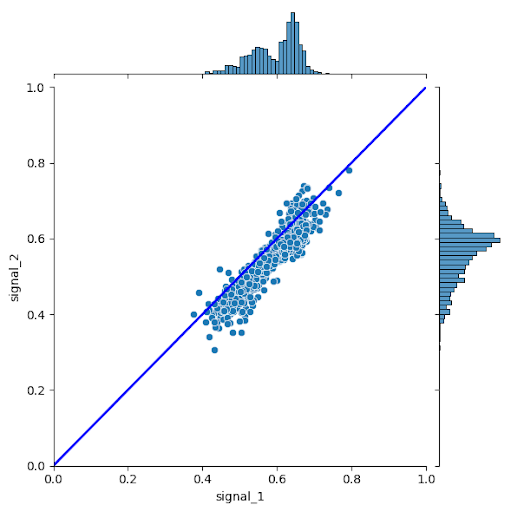}
\end{subfigure}
\hspace{-14pt}
\begin{subfigure}{0.29\linewidth}
  \includegraphics[width=\linewidth]{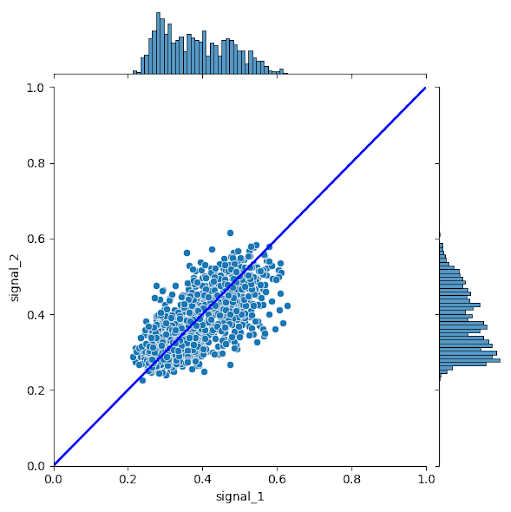}
\end{subfigure}
\hspace{-5pt}
\begin{subfigure}{0.29\linewidth}
  \includegraphics[width=\linewidth]{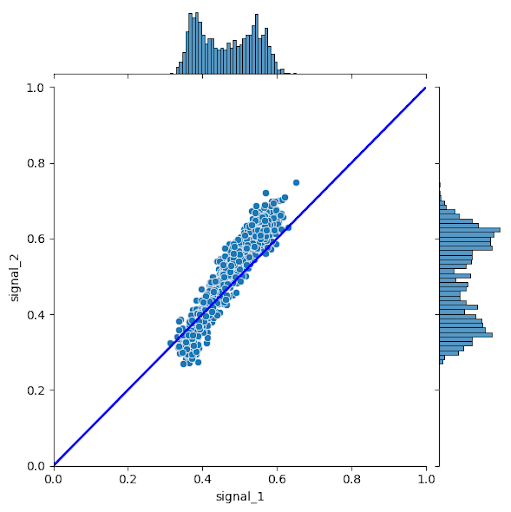}
\end{subfigure}

\bigskip

Without CD
\begin{subfigure}{0.29\linewidth}
  \includegraphics[width=\linewidth]{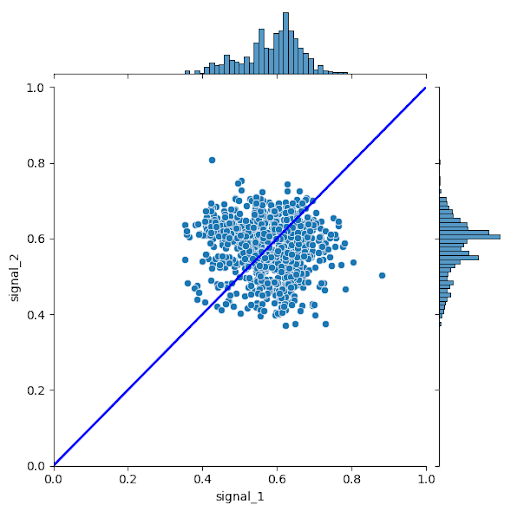}
  \caption{Simple Sine} \label{fig:WD_Simple_Sine}
\end{subfigure}
\hspace{-10pt}
\begin{subfigure}{0.29\linewidth}
  \includegraphics[width=\linewidth]{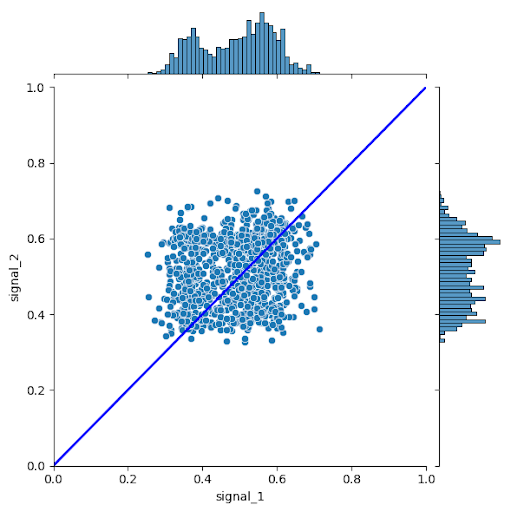}
  \caption{Sine with frequency changes} \label{fig:WD_freq_change}
\end{subfigure}
\hspace{-5pt}
\begin{subfigure}{0.29\linewidth}
  \includegraphics[width=\linewidth]{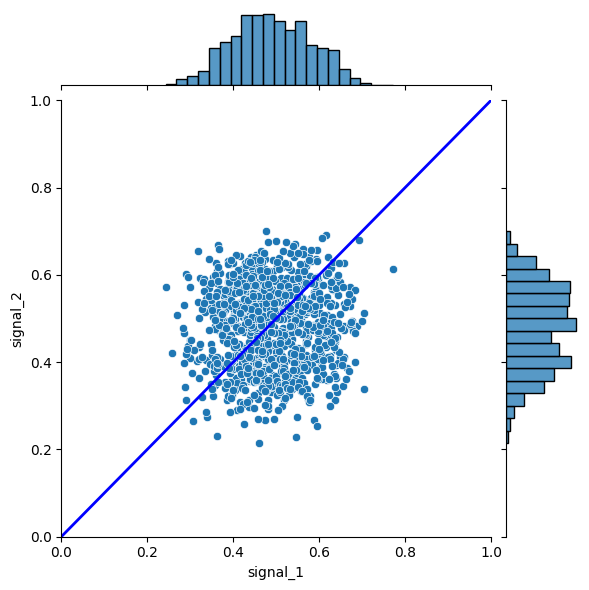}
  \caption{Sine with anomaly} \label{fig:WD_anomaly}
\end{subfigure}
\caption{\centering Amplitudes from generated signals. This diagram depicts the trade-off between Diversity (AWD) and Correlation Analysis (AED). Each point represents the signal amplitude in the first channel (x-axis) vs the signal amplitude in the second channel (y-axis). The marginal distributions of the amplitudes in the corresponding channel are shown on the side of each figure.}
\label{fig:wdvsmse}
\vspace{-20pt}
\end{figure*}

Figure \ref{fig:wdvsmse} shows the qualitative results of the experiment described in Section \ref{sec:Toy-diversity}. The information in Table \ref{table:wd-table} was derived from these plots. As previously discussed, the amplitudes of two channels of data generated with COSCI-GAN with Central Discriminator are more similar to each other than in the absence of a Central Discriminator. This similarity, however, comes at the expense of losing similarity between the generated time-series' marginal distribution of amplitudes and the toy Simple Sine time-series' marginal distribution of amplitudes.

\subsection{Toy Sine Datasets - Feature-based Correlation Analysis}
\label{sec:appen-Toy-correlation}

\begin{figure}[H]
\vspace{-5pt}
  \centering
  \includegraphics[width=\linewidth]{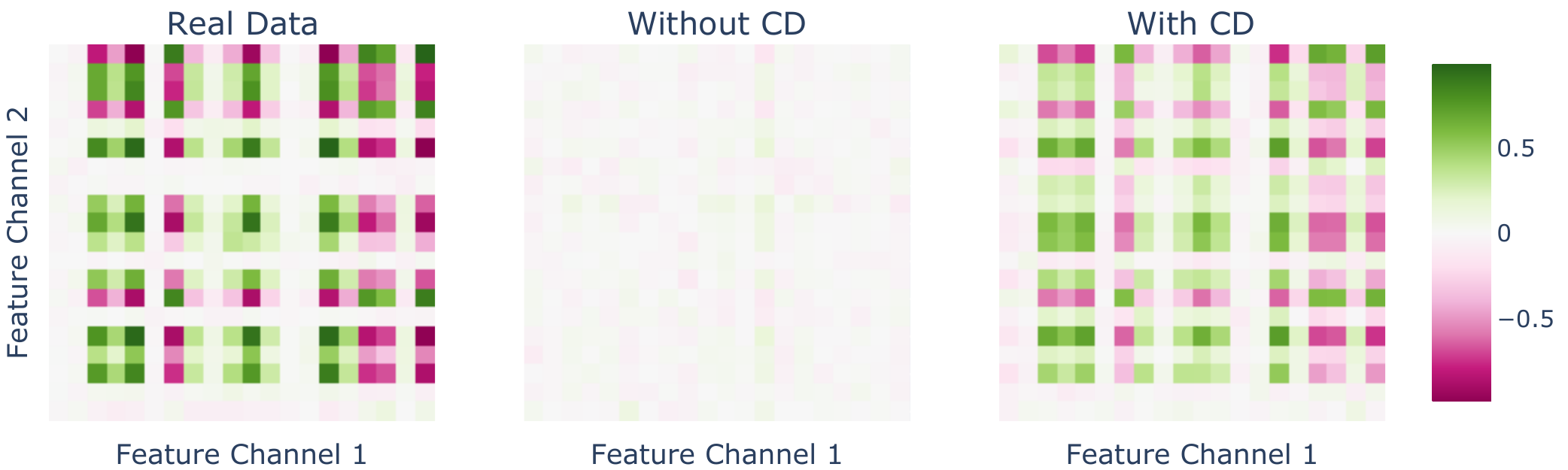}
  \caption{Heatmaps for Catch22 Pairwise Correlations for Sine with Frequency Changes.}
  \label{fig:correlation_freq_change}
\end{figure}

\begin{figure}[H]
\vspace{-5pt}
  \centering
  \includegraphics[width=\linewidth]{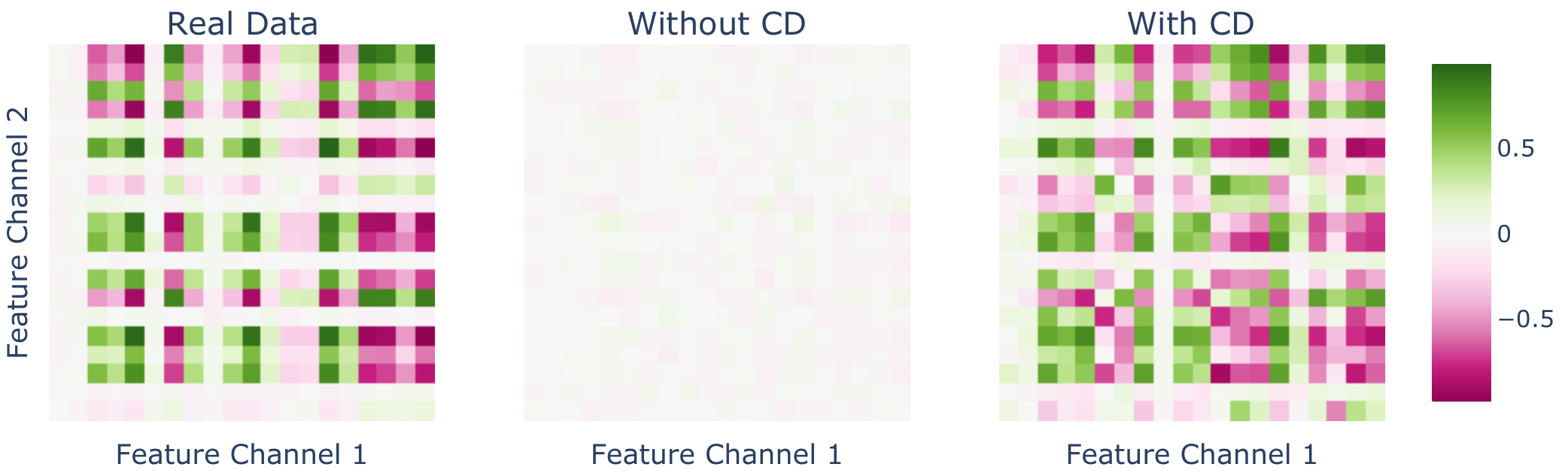}
  \caption{Heatmaps for Catch22 Pairwise Correlations for Sine with Anomaly.}
  \label{fig:correlation_anomaly}
\end{figure}

We computed the same heatmap as Figure \ref{fig:correlation} for the other two toy datasets, as mentioned in Section \ref{sec:correlation-analysis}. Figure \ref{fig:correlation_freq_change} depicts a heatmap for the Sine with frequency changes dataset, while Figure \ref{fig:correlation_anomaly} represents a heatmap for the Sine with anomaly dataset. When we generate fake time-series, we are looking for the preservation of correlation patterns. As shown in the heatmaps, and as demonstrated in Section \ref{sec:correlation-analysis}, COSCI-GAN without a Central Discriminator cannot preserve correlation patterns in real data as good as COSCI-GAN with a Central Discriminator.


\subsection{EEG Eye State Dataset and Downstream Classification}

\subsubsection{Classifier Structure}
\label{sec:appen-EEG-classifier}

\begin{figure}[H]
  \centering
  \includegraphics[width=0.6\linewidth]{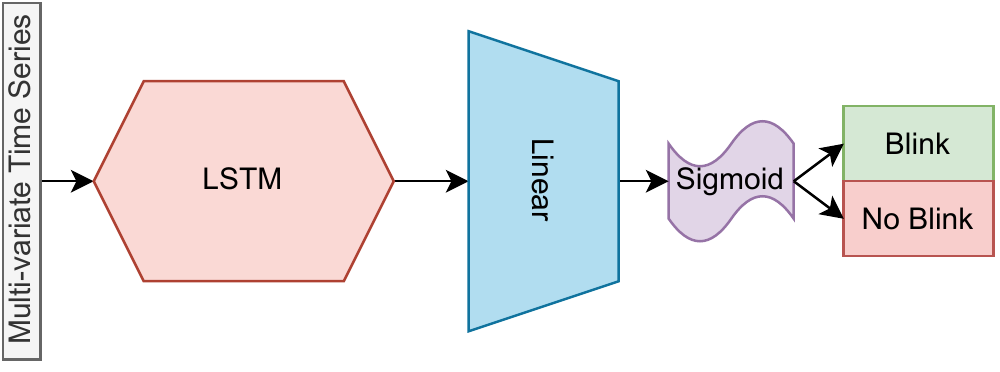}
  \caption{Structure of the blink classifier.}
  \label{fig:classifier}
\end{figure}

We designed a classification task on an EEG eye state dataset, as described in Section \ref{sec:EEG}. As Figure \ref{fig:classifier} illustrates, the classifier we used for this classification task is a one-layer LSTM module with a hidden dimension of $256$, followed by a linear layer that maps the LSTM module's output to a single number, which is then fed into a Sigmoid function. Our classifier determines whether or not the multi-variate time-series contains a blink based on the final output. In each experiment, we performed cross-validation and trained on the training data ($64\%$ of the real data) to find the best network based on the validation data prediction results ($16\%$ of the real data), and then reported the test data prediction results ($20\%$ of the real data). TensorFlow\cite{tensorflow2015-whitepaper} is used to implement the classifier.

\subsubsection{All-Fake and Augmentation Experiments}
\label{sec:appen-EEG-Experiments}

\begin{figure}[H]\centering
\subfloat[Accuracy of All-Synthetic Experiment.]
{\label{fig:New_All_fake}\includegraphics[width=.49\linewidth]{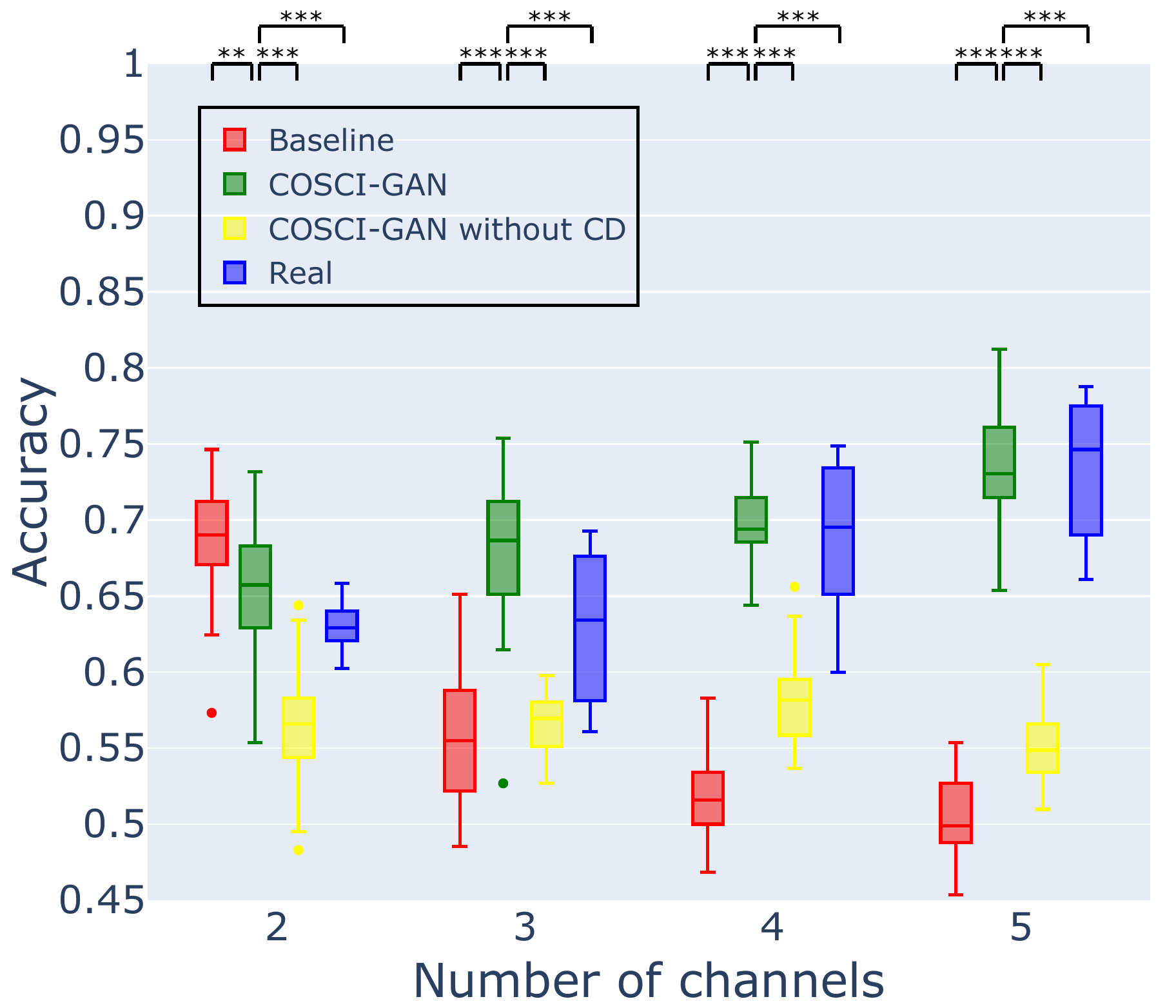}}\hfill
\subfloat[Accuracy of Augmentation Experiment.]
{\label{fig:New_Augmentation}\includegraphics[width=.49\linewidth]{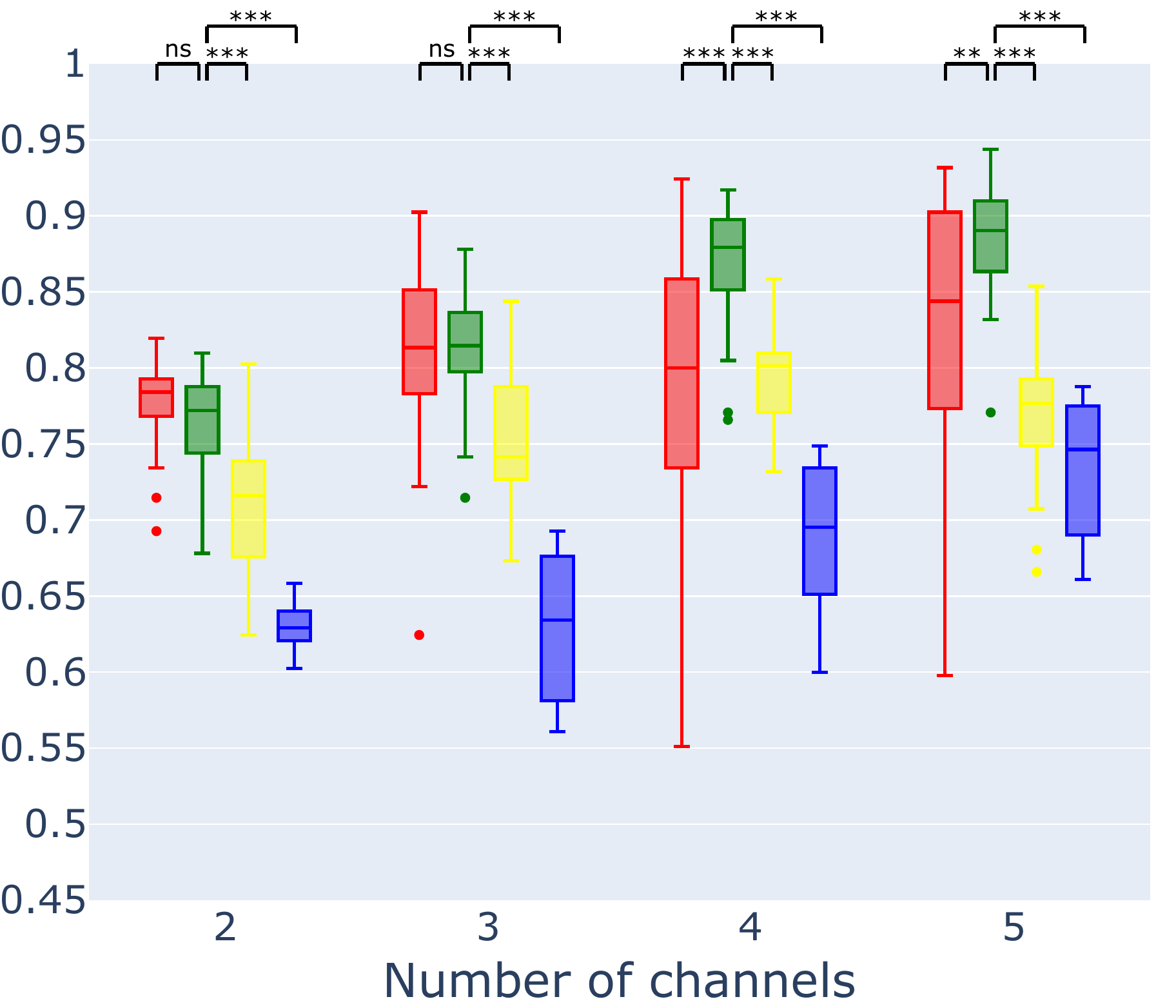}}
\caption{\centering Complete results of experiment three. In figure (a) and (b), T-tests were applied to the results in order to show whether there is a statistically significant difference between the distribution of the results. \textbf{ns} means no significant difference, \textbf{**} means a p-value between $0.01$ and $0.001$, and \textbf{***} means a p-value less than $0.001$.}
\label{fig:complete-allfake-augmentation}
\end{figure}

As described in Section \ref{sec:EEG}, Figure \ref{fig:complete-allfake-augmentation} depicts the complete version of Figures \ref{fig:All_fake} and \ref{fig:Augmentation}, which includes the results of COSCI-GAN without CD. As shown in the figures, removing the Central Discriminators results in even worse accuracy than the baseline method, given that we did not take the channel's correlation into account at all, confirming our hypothesis that preserving channel correlation in a multivariate time-series results in a higher score in downstream tasks.

\subsection{Comparing with State-of-the-art methods on EEG Classification}

\subsubsection{Correlation Analysis of EEG Dataset}
\label{sec:appen-SOTA-Correlation}

\begin{figure}[H]
  \centering
  \includegraphics[width=\linewidth]{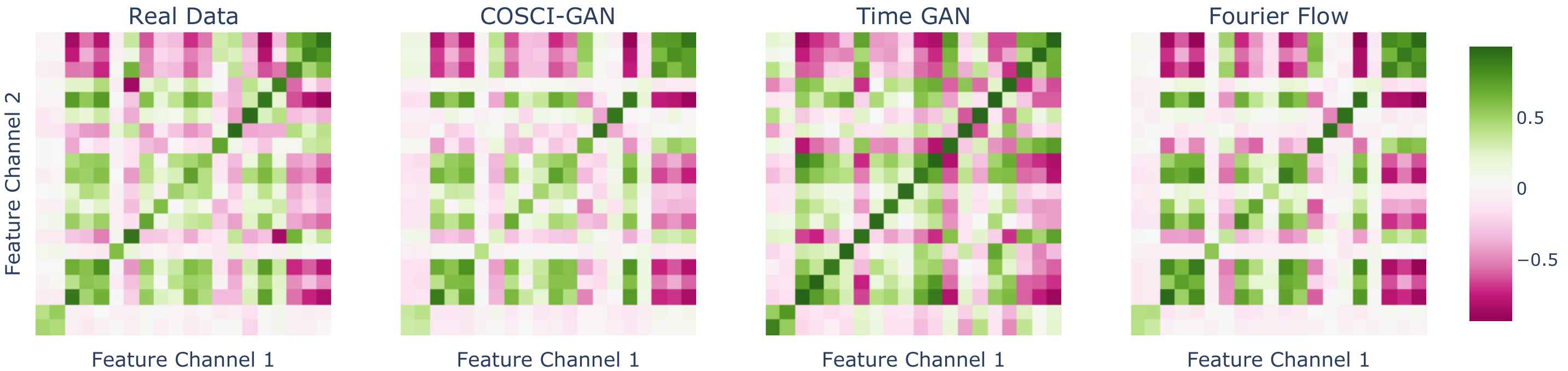}
  \caption{\centering Heatmaps for Catch22 Pairwise Correlations between feature channel 1 and 2 of the EEG eye state dataset.}
  \label{fig:correlation_1_2}
\end{figure}

We repeated the correlation analysis from Section \ref{sec:correlation-analysis} and generated figures similar to Figure \ref{fig:correlation} to compare COSCI-GAN with TimeGAN and Fourier Flows in terms of correlation preservation, as described in Section \ref{sec:sota}. Because there are five channels in the EEG dataset, we will have ten pairs of features and thus ten heatmaps per method. Because the patterns in all ten heatmaps are nearly identical, we only show one of them as an example in Figure \ref{fig:correlation_1_2}. Table \ref{table:correlation-sota} contains a summary of the similarity between the data generated using each method and real dataset for all ten pairs of channels.

\newpage

\subsubsection{PCA and t-SNE Visualization}
\label{sec:appen-SOTA-PCA}

\begin{figure}[H]
PCA
\hspace{2pt}
\begin{subfigure}{0.3\linewidth}
  \includegraphics[width=\linewidth]{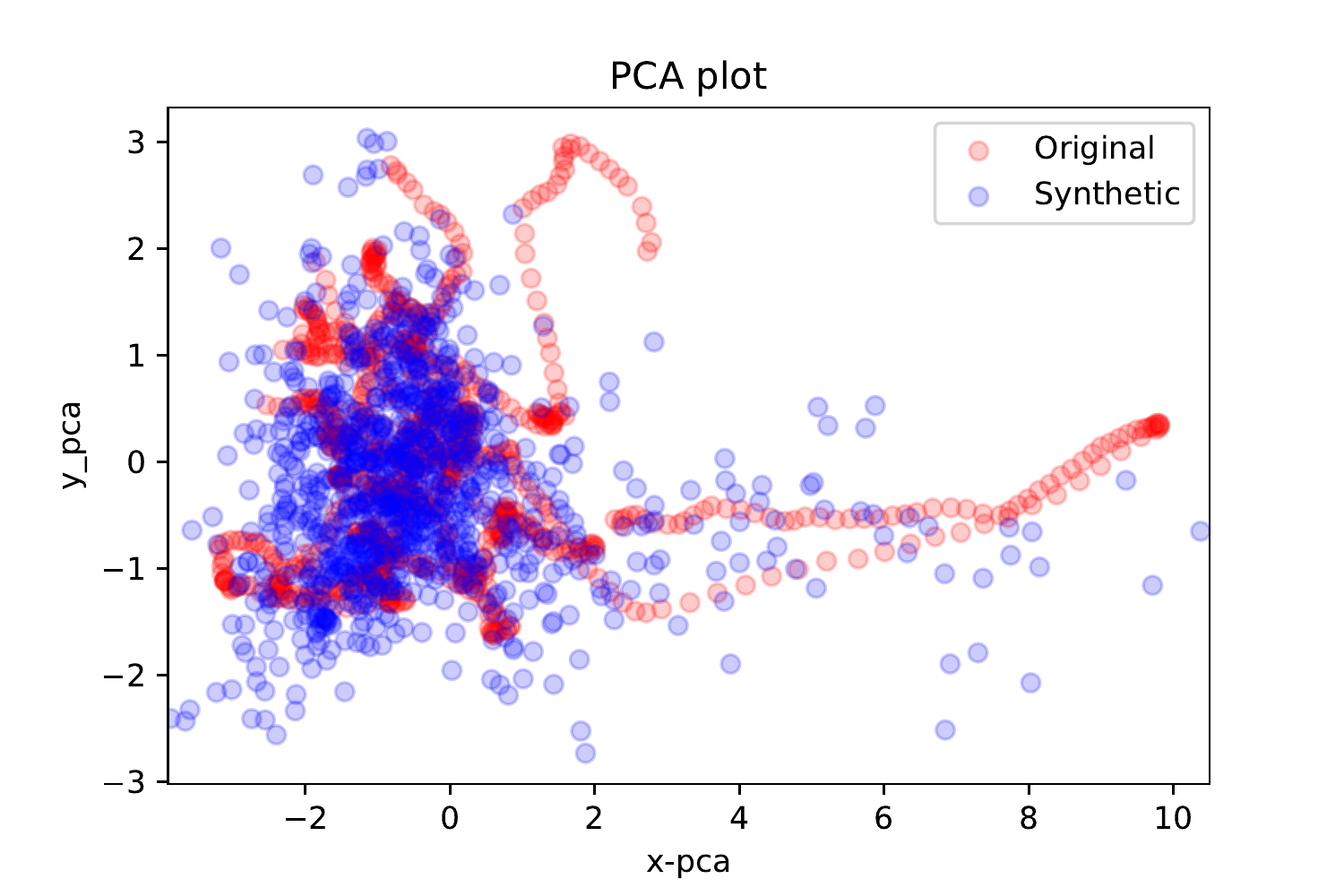}
\end{subfigure}
\hspace{-6pt}
\begin{subfigure}{0.3\linewidth}
  \includegraphics[width=\linewidth]{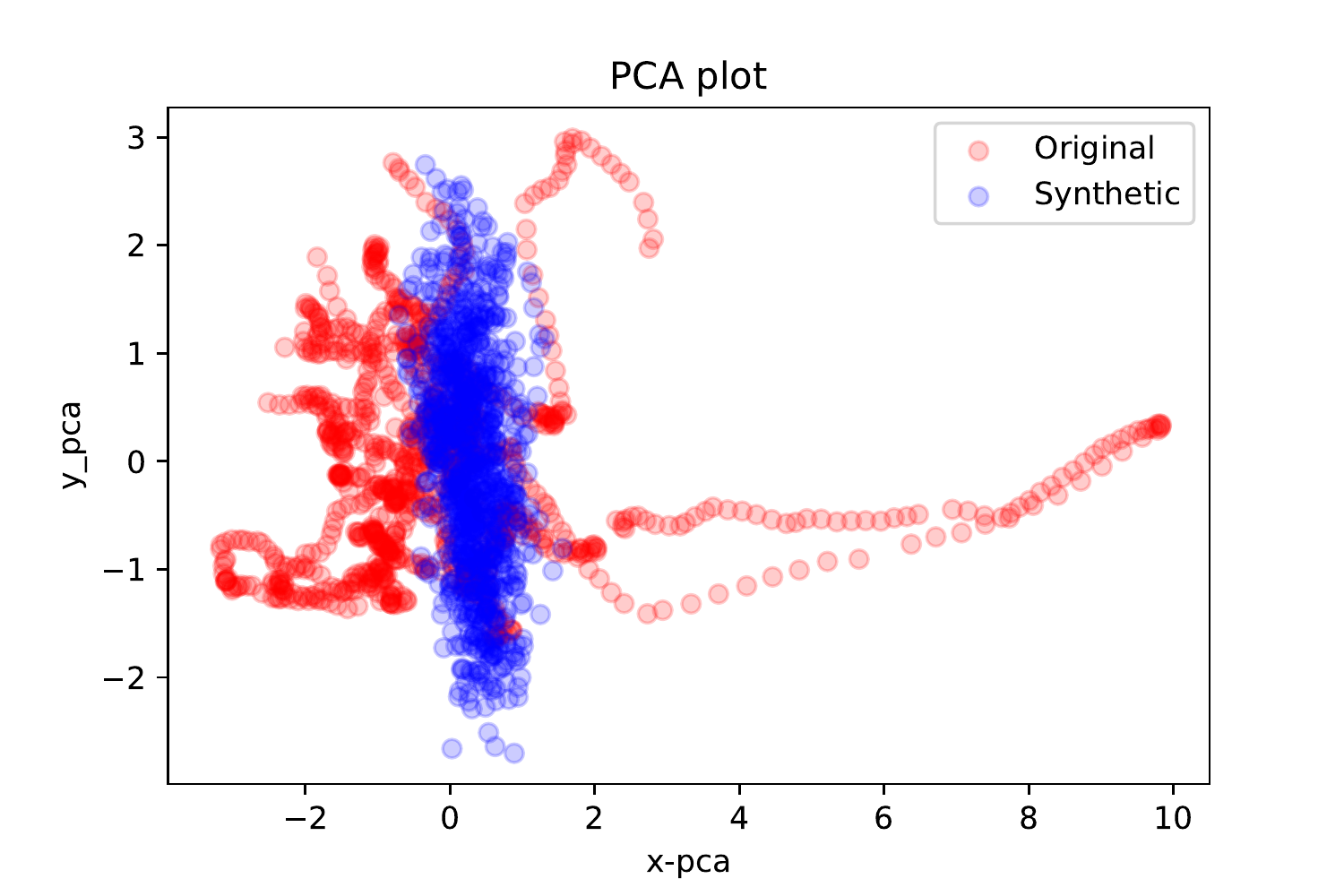}
\end{subfigure}
\hspace{-5pt}
\begin{subfigure}{0.3\linewidth}
  \includegraphics[width=\linewidth]{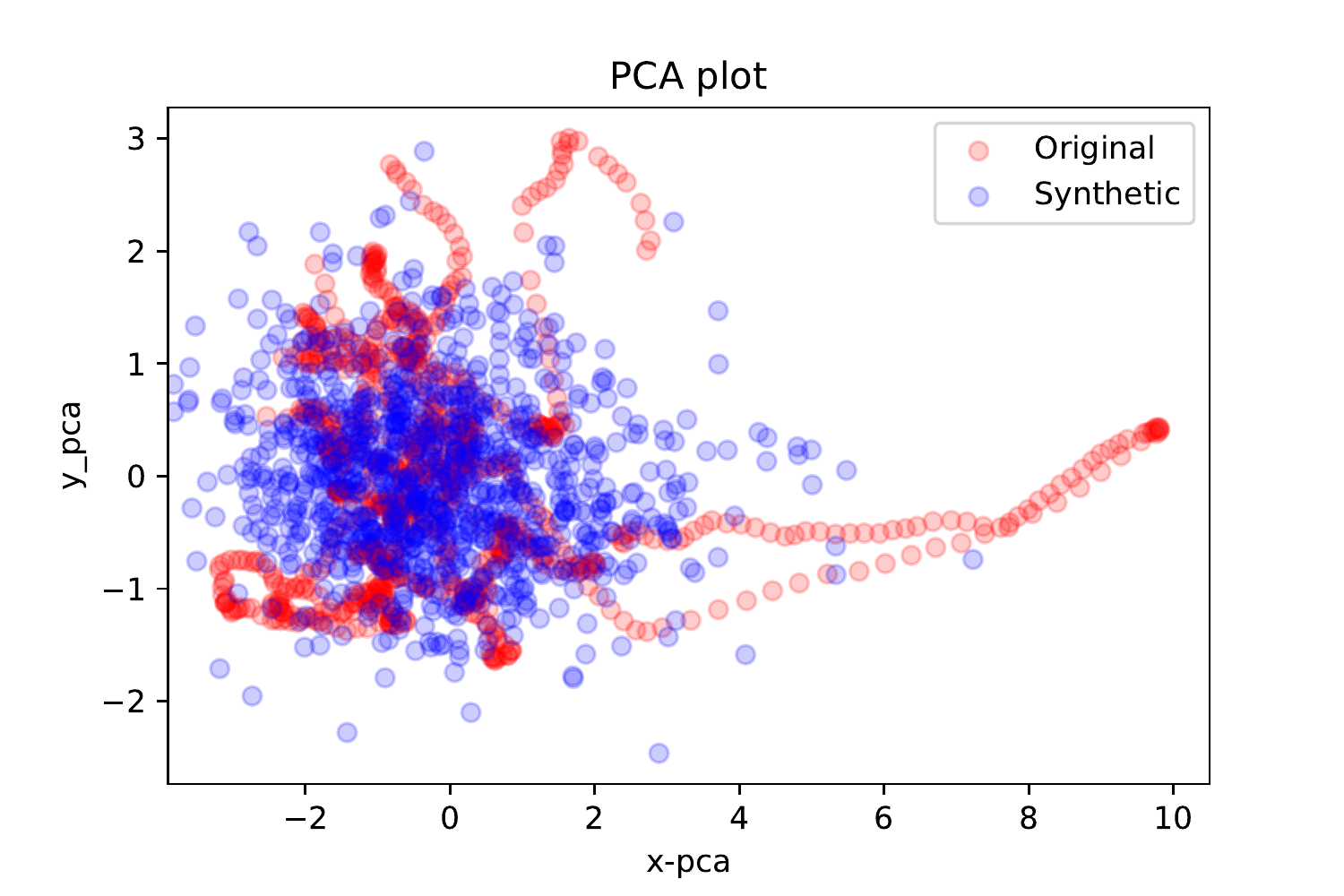}
\end{subfigure}
\bigskip
\\
t-SNE
\hspace{-4pt}
\begin{subfigure}{0.3\linewidth}
  \includegraphics[width=\linewidth]{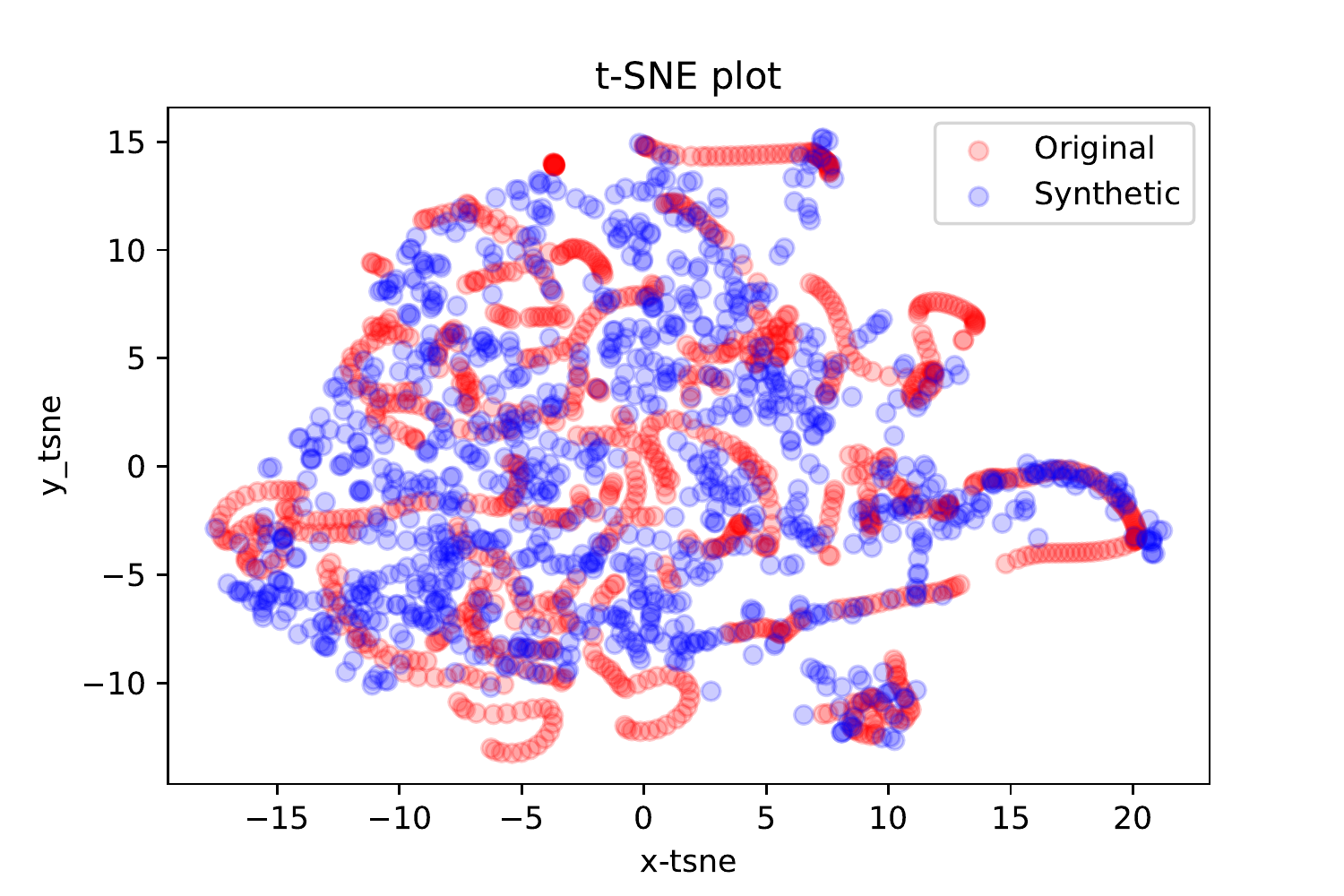}
  \caption{COSCI-GAN} \label{fig:COSCI-GAN_tsne_plot_eeg}
\end{subfigure}
\hspace{-6pt}
\begin{subfigure}{0.3\linewidth}
  \includegraphics[width=\linewidth]{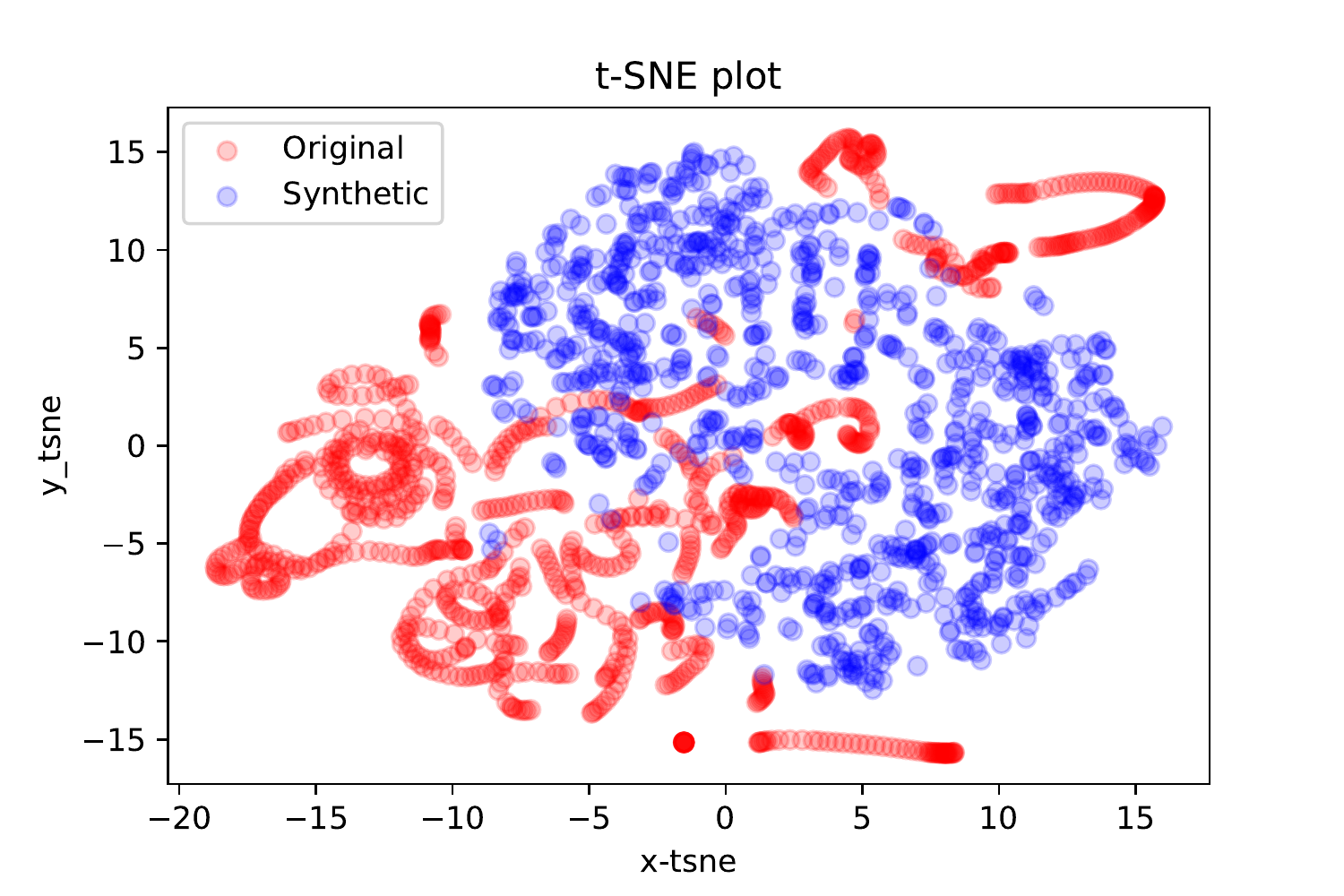}
  \caption{TimeGAN} \label{fig:TimeGAN_tsne_plot_eeg}
\end{subfigure}
\hspace{-5pt}
\begin{subfigure}{0.3\linewidth}
  \includegraphics[width=\linewidth]{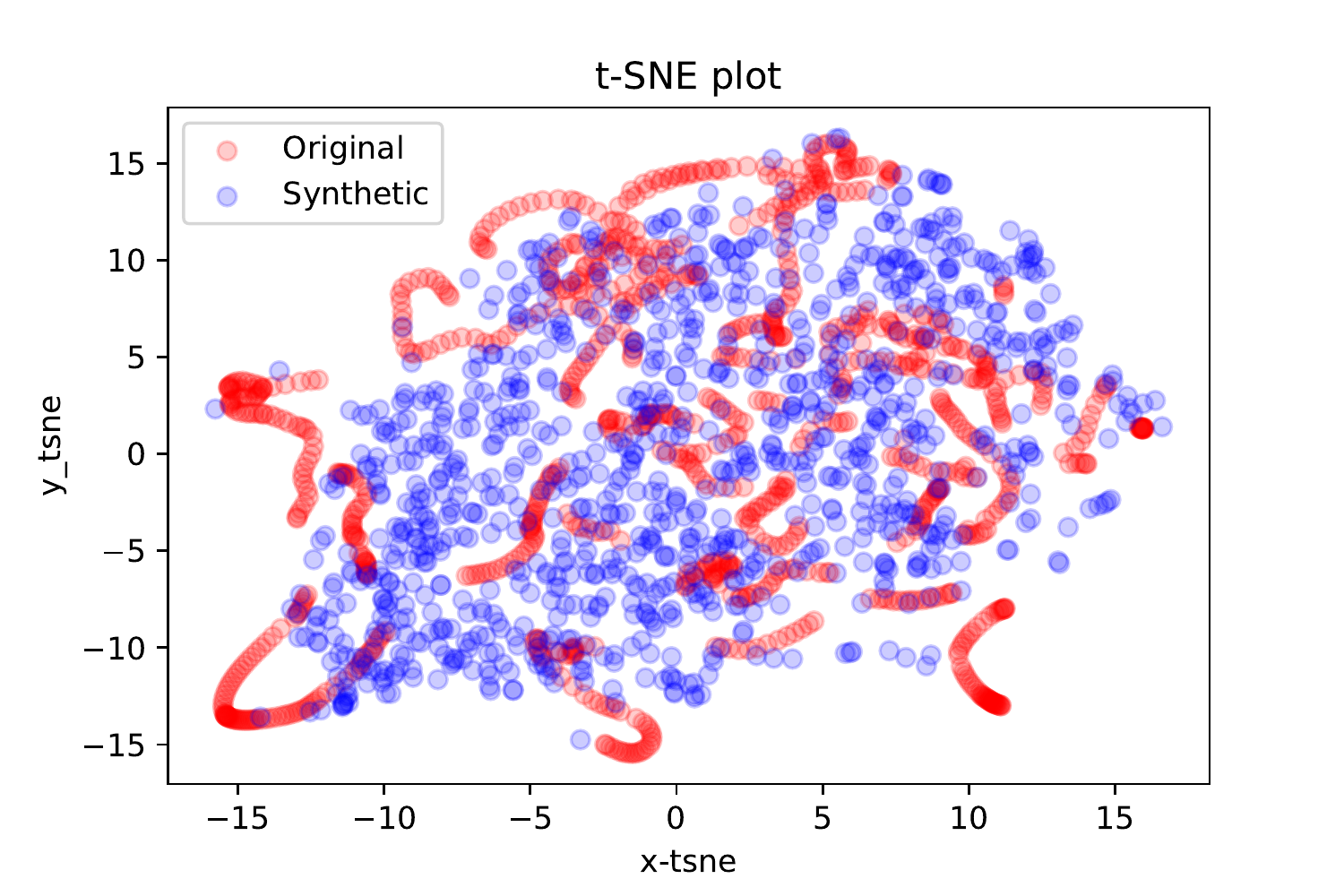}
  \caption{Fourier Flows} \label{fig:FF_tsne_plot_eeg}
\end{subfigure}
\caption{\centering Qualitative Comparison of methods - EEG Dataset. Red dots represent real data instances, and blue dots represent generated data samples in all plots.}
\label{fig:Qualitative_eeg}
\end{figure}

\begin{figure}[H]

PCA
\hspace{2pt}
\begin{subfigure}{0.3\linewidth}
  \includegraphics[width=\linewidth]{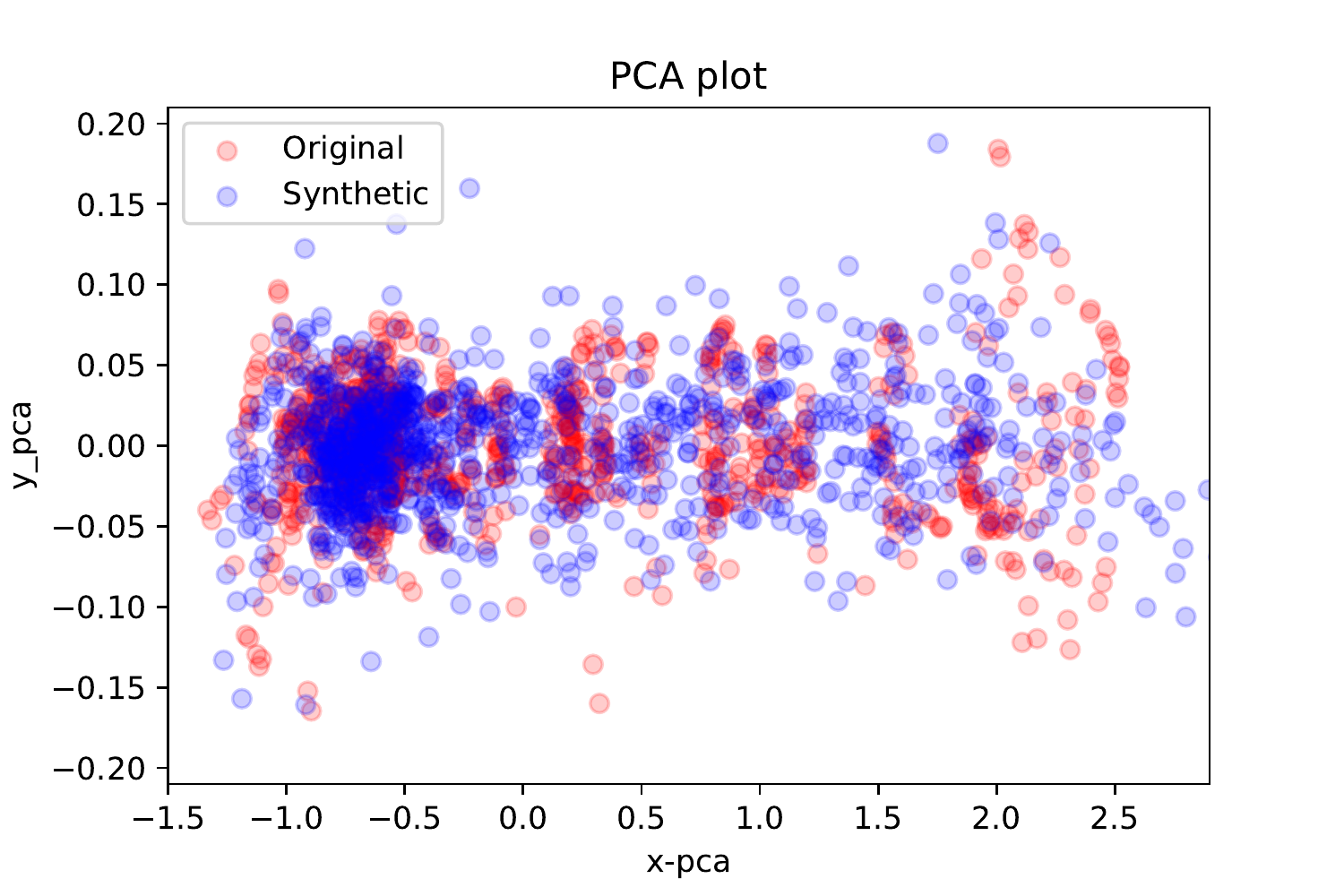}
\end{subfigure}
\hspace{-6pt}
\begin{subfigure}{0.28\linewidth}
  \includegraphics[width=\linewidth]{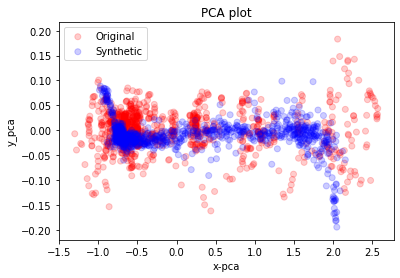}
\end{subfigure}
\begin{subfigure}{0.3\linewidth}
  \includegraphics[width=\linewidth]{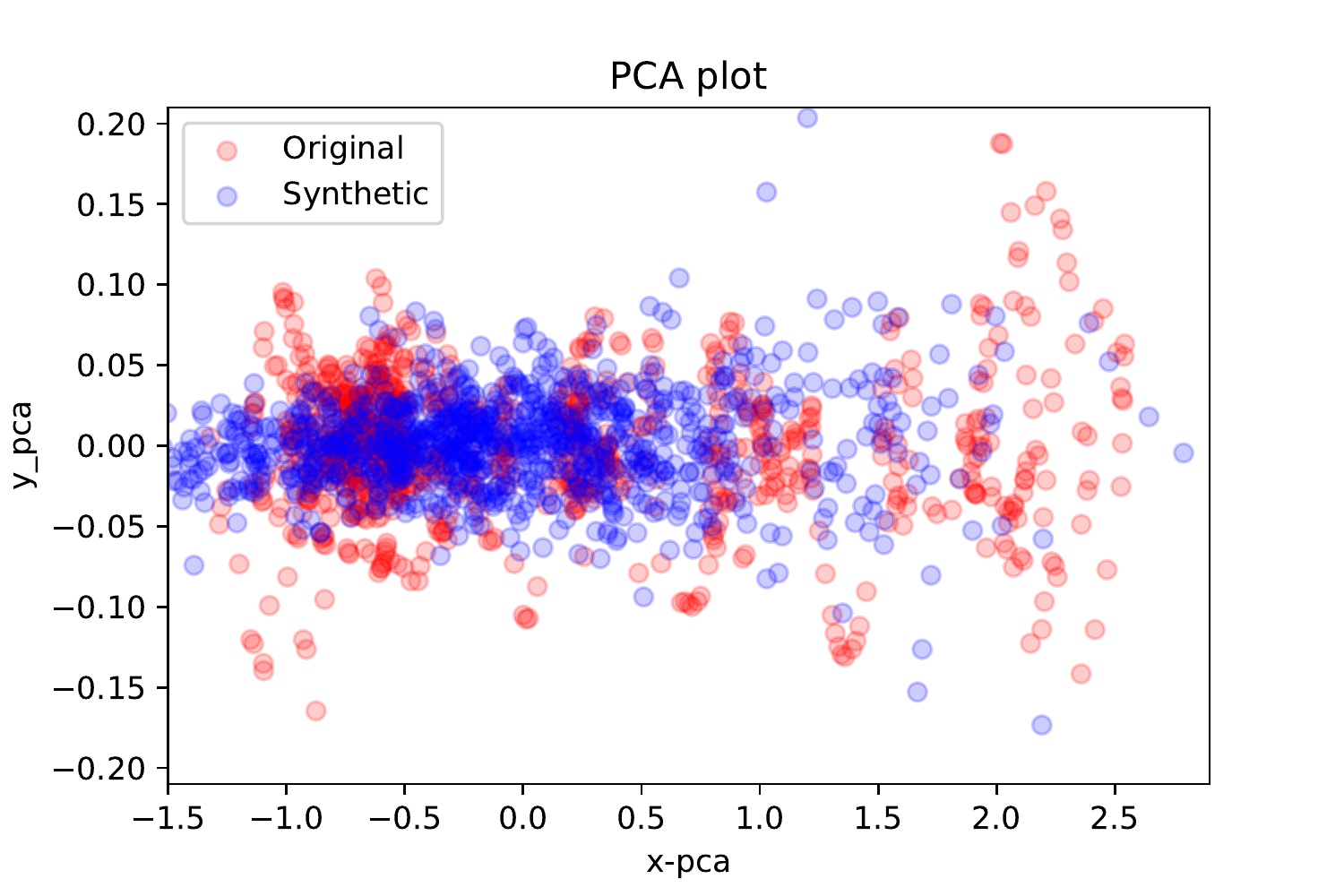}
\end{subfigure}
\bigskip
\\
t-SNE
\hspace{-4pt}
\begin{subfigure}{0.3\linewidth}
  \includegraphics[width=\linewidth]{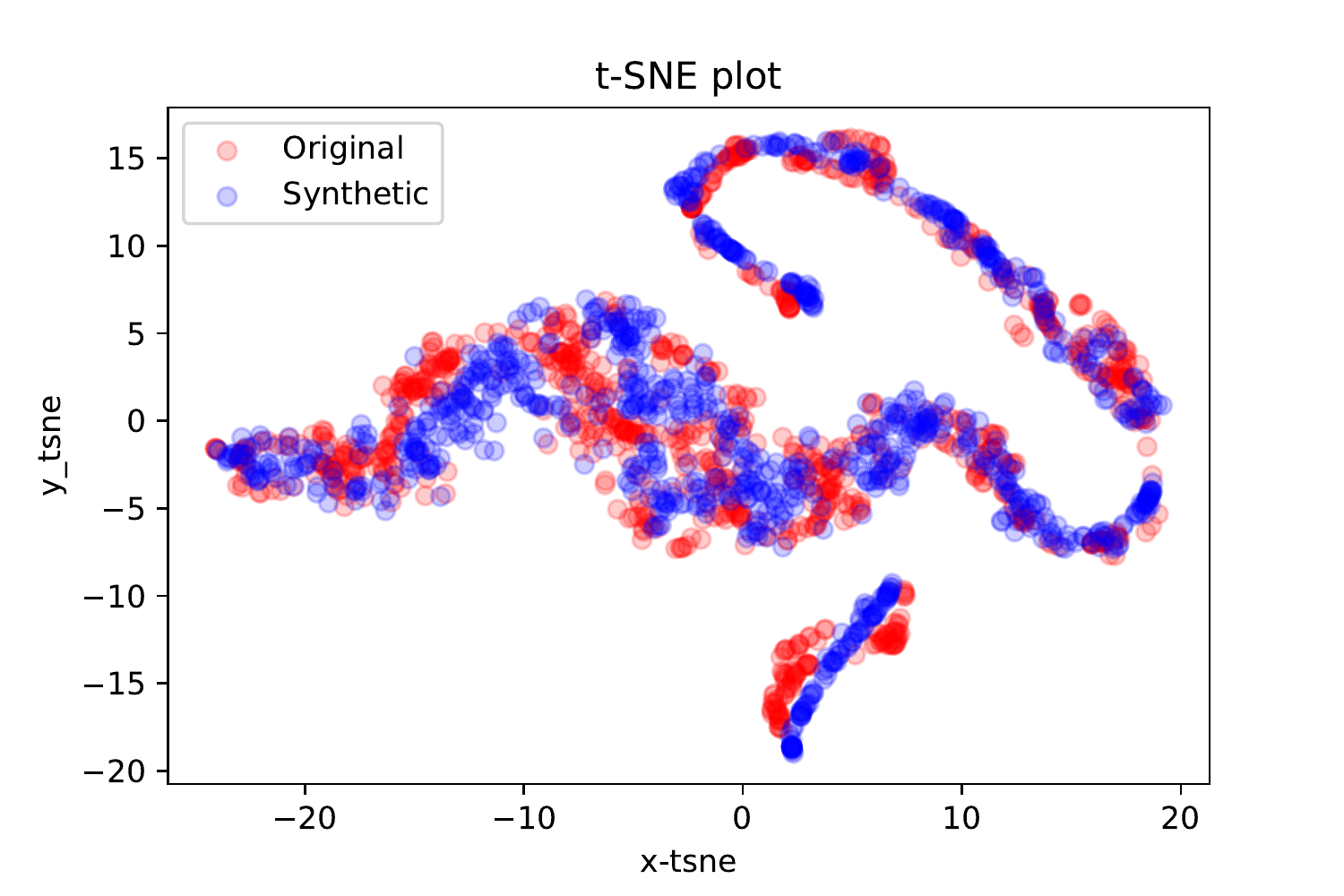}
  \caption{COSCI-GAN} \label{fig:COSCI-GAN_tsne_plot_stock}
\end{subfigure}
\hspace{-6pt}
\begin{subfigure}{0.28\linewidth}
  \includegraphics[width=\linewidth]{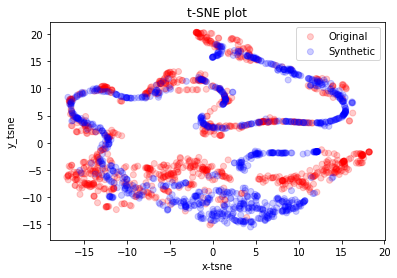}
  \caption{TimeGAN} \label{fig:TimeGAN_tsne_plot_stock}
\end{subfigure}
\begin{subfigure}{0.3\linewidth}
  \includegraphics[width=\linewidth]{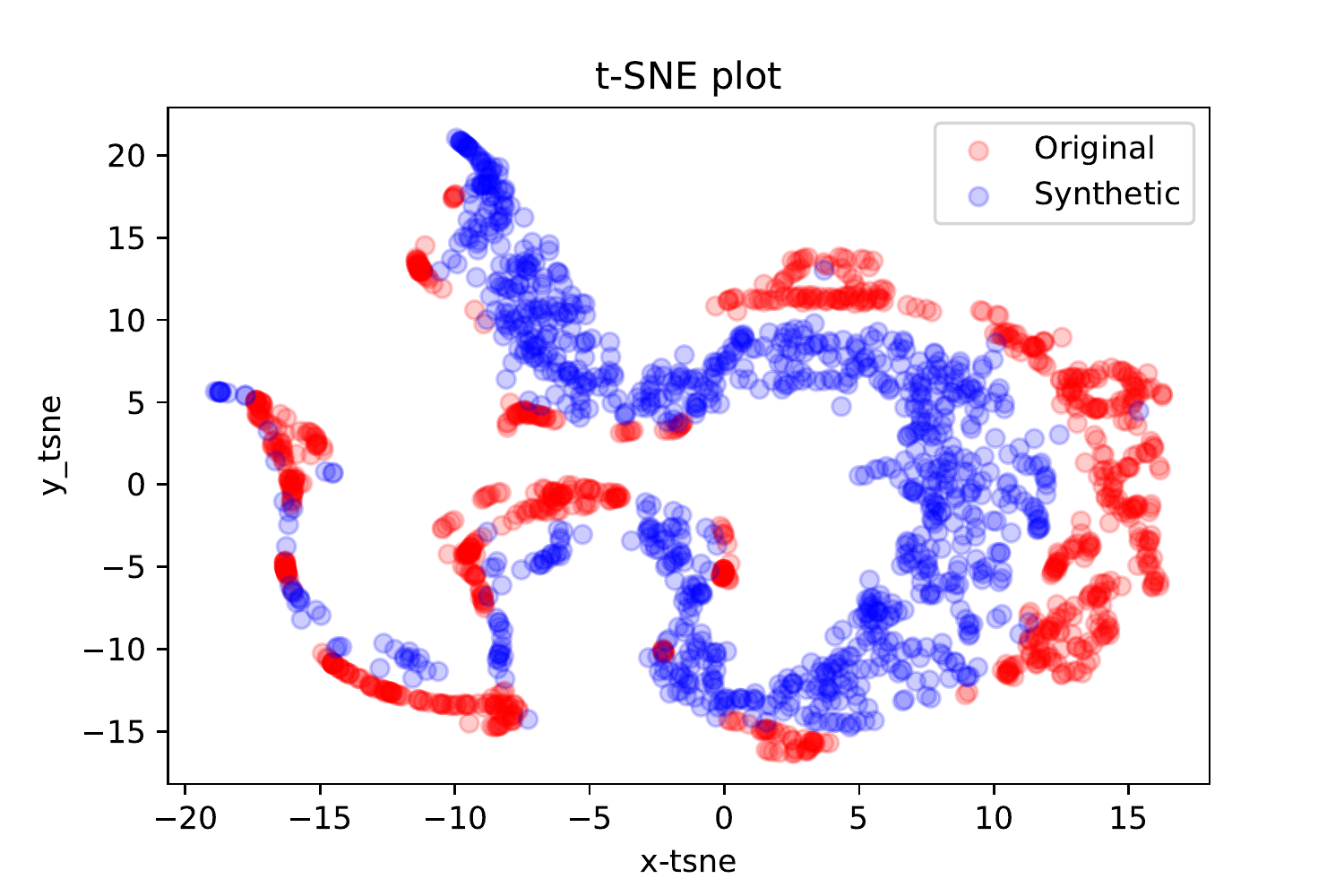}
  \caption{Fourier Flows} \label{fig:FF_tsne_plot_stock}
\end{subfigure}
\caption{\centering Qualitative Comparison of methods - Stock Dataset. Red dots represent real data instances, and blue dots represent generated data samples in all plots.}
\label{fig:Qualitative_stock}
\end{figure}

In order to compare the diversity of the generated time-series using COSCI-GAN, TimeGAN, and Fourier Flows with the real dataset, we used the qualitative analysis described in the last paragraph of Section \ref{sec:sota}. We applied PCA and t-SNE analyses to visualize how well the generated time-series distributions cover the real data distributions in 2-dimensional space. Figure \ref{fig:Qualitative_eeg} shows the results for the EEG eye state dataset, and Figure \ref{fig:Qualitative_stock} shows the results for the Stock dataset. According to the figures, synthetic datasets generated by COSCI-GAN have significantly more overlap with the original data than other SOTA methods. Red dots represent real data instances, and blue dots represent generated data samples in all plots.

\end{appendix}

\end{document}